\documentclass{article}

\usepackage{arxiv}

\usepackage[utf8]{inputenc} 
\usepackage[T1]{fontenc}    
\usepackage{hyperref}       
\usepackage{url}            
\usepackage{booktabs}       
\usepackage{amsfonts}       
\usepackage{nicefrac}       
\usepackage{microtype}      
\usepackage{lipsum}

\usepackage{lineno}
\usepackage{graphicx}
\usepackage{subfigure}
\usepackage{array}
\usepackage{amssymb}
\usepackage{amsmath}
\usepackage{algorithm}
\usepackage{algorithmic}

\title{Deep Clustering with a Dynamic Autoencoder: From Reconstruction towards Centroids Construction}

\author{
    $\: \: \: \: \: \:$ Nairouz Mrabah\\
    $\: \: \: \: \: \:$  National Engineering School of Tunis\\
    $\: \: \: \: \: \:$  University of Tunis El Manar\\
    $\: \: \: \: \: \:$ Tunis, Tunisia\\
  $\: \: \: \: \: \:$ \texttt{nairouz.mrabah@ensi-uma.tn} \\
  \And
  Naimul Mefraz Khan\\
  Department of Electrical and Computer Engineering\\
  Ryerson University\\
  Toronto, ON, Canada\\
  \texttt{n77khan@ee.ryerson.ca} \\
  \And
  Riadh Ksantini\\
  Higher School of Communication of Tunis\\
  University of Carthage\\
  Tunis, Tunisia\\
  \texttt{riadh.ksantini@supcom.tn}\\
  \And
  Zied Lachiri\\
  National Engineering School of Tunis\\
  University of Tunis El Manar\\
  Tunis, Tunisia\\
  \texttt{zied.lachiri@enit.rnu.tn}\\
}

\begin{document}
\maketitle

\begin{abstract}
In unsupervised learning, there is no apparent straightforward cost function that can capture the significant factors of variations and similarities. Since natural systems have smooth dynamics, an opportunity is lost if an unsupervised objective function remains static during the training process. The absence of concrete supervision suggests that smooth dynamics should be integrated. Compared to classical static cost functions, dynamic objective functions allow to better make use of the gradual and uncertain knowledge acquired through pseudo-supervision. In this paper, we propose Dynamic Autoencoder (DynAE), a novel model for deep clustering that overcomes a clustering-reconstruction trade-off, by gradually and smoothly eliminating the reconstruction objective function in favor of a construction one. Experimental evaluations on benchmark datasets show that our approach achieves state-of-the-art results compared to the most relevant deep clustering methods. \textit{\href{https://github.com/nairouz/DynAE}{github.com/nairouz/DynAE}} 
\end{abstract}


\section{Introduction}

Clustering has been studied for several decades. Multiple approaches have been proposed to partition the data according to intrinsic similarities and without using significant supervisory signals. However, the clustering problem remains an open question for three main reasons:
\begin{enumerate}
  \item \textit{Large-scale/big data} \cite{paper57}, resulting in high computational complexity. Dealing with this problem requires processing the excessive amounts within an acceptable time. 
  \item \textit{High-dimensional} data. For instance, the average size (width $\times$ height) on ImageNet \cite{paper82} exceeds 181000 pixels. This problem is well-known to the scientific community as the curse of dimensionality \cite{paper56}. High-dimensional data means a high-dimensional space. Increasing the space volume leads to data sparsity, which is problematic for statistical significance. Furthermore, a high number of features increases the chances of overfitting.
  \item  \textit{High-semantic} data is one of the most challenging problems. Data with high-level semanticity (human level), such as, speech, textual data, images, and videos, generally have complex similarity patterns, which are difficult to pin down mathematically. Two images belonging to the same class (semantically) can have completely different pixel representations. However, it is well-known that high-semantic data is hierarchical by nature \cite{paper20}. It means that high-level features are composed of lower-level ones. In the case of image datasets, objects are a combination of motifs, which in turn are a combination of edges. This compositional aspect generalizes to speech and text. We utilize this hierarchical nature of semanticity as a motivation. 
\end{enumerate}

We can classify the existing clustering algorithms into four main categories according to their contributions in resolving the three main challenges (i.e., high-dimensional, large-scale, high-semantic data). a) classical clustering  approaches, b) subspace clustering approaches, c) manifold clustering approaches and d) deep clustering approaches.

Classical clustering approaches like K-Means \cite{paper14}, DBSCAN \cite{paper83} and Agglomerative Clustering (AC) \cite{paper84} capture similarities based on a notion of distance in the original data space. As a result, they are considered to be shallow models. Although being successfully used in multiple applications, shallow models fall short of reliable discriminative abilities. Put it differently, computing distance-based metrics in the raw data-space is not sufficient for discovering semantic similarities. Added to that, standard implementations of K-Means, DBSCAN and Hierarchical clustering require $o(n^{2})$ time and $o(n^{2})$ memory, where $n$ is the number of samples. This quadratic complexity makes the processing of large-scale datasets prohibitive. Furthermore, the shallow models do not overcome the curse of dimensionality challenge, since the data separation is performed in the raw data space.

In the interest of evading the curse of dimensionality, a bunch of dimensionality reduction techniques have been proposed. Linear transformations like Principal Component Analysis (PCA) \cite{paper49} and Factor Analysis \cite{paper50} are broadly adopted for a multitude of applications. Existing non-linear techniques rely on the manifold assumption \cite{paper52}, which postulates the existence of low-dimensional manifolds, where the embedded information is concentrated. Multi-dimensional scaling (MDS) \cite{paper51} and Isometric Feature Mapping (Isomap) \cite{paper52} are among the most commonly used non-linear dimensionality reduction approaches. While these methods target capturing the most relevant information, they are subject to discriminative information loss, which deteriorates the clustering performance. 

To tackle the curse of dimensionality while prioritizing categorical separability, subspace clustering \cite{paper13} enables to find relevant dimensions spanning a subspace for each cluster. Contrary to typical dimensionality reduction techniques, subspace clustering does not overlook the discriminative information available in data. However, they are computationally costly. Besides, the existence of clusters within linear subspaces does not conform to the complexity of natural datasets.

Other than subspace clustering, manifold clustering \cite{paper12} also combines discriminative dimensionality reduction with clustering. However, instead of confining the data to low dimensional subspaces, manifold clustering maps the samples to non-linear manifolds by computing a similarity matrix.
Spectral K-Means and spectral clustering are popular manifold clustering approaches. Under mild conditions, spectral clustering can be seen as a kernel K-Means problem \cite{paper97}. The computational time of this family grows considerably for large-scale datasets. Moreover, the representation power of such methods is limited. The discriminative ability of the similarity matrix usually underfits the natural data semanticity.

Recently, deep learning has shown great promise in solving pattern recognition problems \cite{paper85, paper86}. It allows to learn space transformations and gradually extract higher semantic representations from one layer to another. The immense success of deep learning is mostly imputed to the convenience of using multi-layers architectures in solving data-oriented problems. In practice, three principal reasons can mostly explain the suitability of these models. First, mini-batch Stochastic Gradient Descent (SGD) grants neural networks the ability to elude the time and memory bottlenecks. Thus, multi-layers architecture can be used to process large-scale databases. Second, neural networks are attractive for high-dimensional data since it is possible to project the samples in a lower-dimensional space. Third, the compositional aspect of natural datasets plays well with the hierarchical architecture of a deep neural network. Therefore, multi-layers architecture is suitable for processing high-semantic data. Despite all these advantages, discovering hidden data structures without leveraging any supervisory signal remains an open and challenging research area.

Most of the existing deep clustering methods harness an auto-encoding architecture \cite{paper27,paper28,paper29,paper30,paper31}. Some other approaches rely on an encoding network without a decoder \cite{paper21,paper22,paper23,paper24,paper26}. The latter strategy consists of two alternative stages: grouping the embedded features based on a typical clustering algorithm (e.g., K-Means) and updating the encoding weights using the subsequent pseudo-labels as a supervisory signal. For the first strategy, the encoder training is regularized by a reconstruction cost. Hence, the decoder is used to alleviate the effect of training with pseudo-labels. However, eliminating the decoding network and training the encoder based on \textit{hypothetical} similarities can easily corrupt the space topology. Thus, it is possible that the encoder generates \textit{random} discriminative features, which do not reflect the real discriminative characteristics. This problem is called Feature Randomness \cite{paper98}.
 
To mitigate the effect of Feature Randomness, autoencoders are provided with their reconstruction capability, which captures prominent features without forcing any randomness. However, combining clustering and reconstruction is problematic due to the natural trade-off between them. In fact, while the clustering objective function aims to destroy non-discriminative details, the reconstruction objective function allows to preserve all information. Put it differently, all pixels have the same importance in the reconstruction cost regardless of their discriminative role. For instance, a picture of a plane and another of a bird both would contain many blue pixels and few ones representing the real objects. In such a case, the reconstruction would foster encoding information, which is irrelevant for clustering (i.e., the color of the sky). This problem is called Feature Drift \cite{paper98}.

To deal with Feature Randomness and Feature Drift, our proposed model has a dynamic loss function that allows to gradually and smoothly dispense with the reconstruction loss in favor of a construction one. Our experimental results show the suitability of our model when compared to the state-of-the-art clustering methods in terms of accuracy (ACC) and normalized mutual information (NMI). Furthermore, DynAE has reasonable dynamic hyperparameters, which are updated automatically following the dynamics of the learning system. The automatic update of the hyperparameters allows to circumvent cross-validation, which is impractical for unsupervised learning. The contributions of this work are as follows:

\begin{itemize}
  \item Elaborating a novel deep clustering framework, namely, DynAE, which alleviates Feature Randomness.
  \item Using a dynamic loss function to overcome the clustering-reconstruction trade-off.
  \item An improved way of obtaining a K-Means friendly latent space compared to DCN \cite{paper29}.
  \item Establishing a better trade-off between Feature Randomness and Feature Drift compared to ADEC \cite{paper98}, by gradually and smoothly transforming a self-supervised objective into a pseudo-supervised one.
  \item Achieving state-of-the-art clustering performance on real benchmark datasets.
\end{itemize}

\section{Related work}
Most of the existing deep clustering strategies share two simple concepts. The first concept is that deep embedded representations are favorable to clustering. The second concept is that clustering assignments can be used as a supervisory signal to learn embedded representations. Based on that, the existing deep clustering methods can be classified into two main families. 

\begin{figure}[ht]
\vskip 0.2in
\begin{center}
\centerline{\includegraphics[width=300pt, height=125pt]{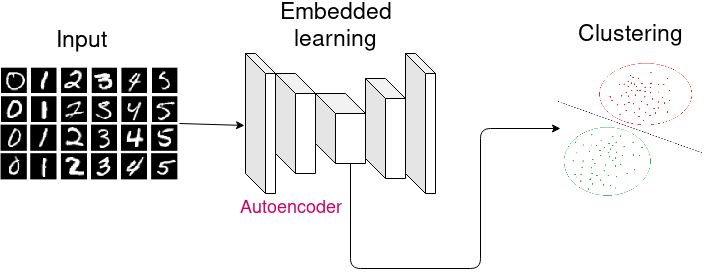}}
\caption{Illustration of the first deep clustering family: embedded learning is performed before clustering.}
\label{fig:deep_clustering_family1}
\end{center}
\vskip -0.2in
\end{figure}

For the first family, embedded learning and clustering are set apart. Particularly, embedded learning precedes clustering. However, the new obtained labels, after the clustering stage, are not used to learn better discriminative features. For instance, in \cite{paper88}, the data is first projected in a lower-dimensional space using an autoencoder, then K-Means clusters the embedded data. Some other interesting research studies revolve around this two-step strategy \cite{paper89, paper90}. In Figure \ref{fig:deep_clustering_family1}, we illustrate the general framework of this family. 

\begin{figure}[ht]
\vskip 0.2in
\begin{center}
\centerline{\includegraphics[width=300pt, height=160pt]{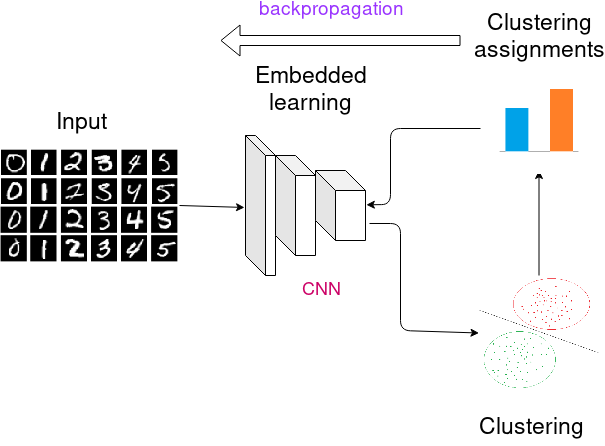}}
\caption{Illustration of the second deep clustering family: embedded learning and clustering are performed jointly.}
\label{fig:deep_clustering_family2}
\end{center}
\vskip -0.2in
\end{figure}

For the second family, embedded learning and clustering are performed jointly. A number of approaches fall under the scope of this group \cite{paper28, paper27, paper29, paper35, paper36, paper24}. Among the common strategies, a classical clustering algorithm is applied to group the embedded data points into different clusters. After that, the encoding transformation is trained to push the embedded space to have a clustering-friendly representation based on the previously obtained clustering assignment. These two steps can be applied alternatively for a specific number of iterations or until meeting a convergence criterion. Figure \ref{fig:deep_clustering_family2} illustrates the joint approach. 

Haeusser et al. \cite{paper21} have proposed to \textit{freeze} the first layers of famous models like VGG \cite{paper91} and Residual networks \cite{paper92} and build the clustering model on top of these layers. Although not using any label from the target training set, this approach strongly relies on features that are extracted in a supervised way (labels from other datasets). Since the low semantic features (e.g., edge detection) are similar across different datasets, we consider such a method to be out of the pure unsupervised learning realm.

Good results have been reported for approaches that heavily depend on strong assumptions for formulating the objective function (e.g., prior knowledge on the size of each cluster) \cite{paper36, paper30, paper73}. Such assumptions may have a considerable influence in clustering ambiguous samples. While these methods may perform well in specific scenarios, the solution is not general enough to be applicable to any dataset.  

Deep Embedding for Clustering (DEC) \cite{paper34} projects the data points in a lower-dimensional space using an autoencoder. After that, the decoder is discarded and the encoder is trained to jointly improve the embedded representations and the clustering centers. DEC requires a pretraining phase before starting to cluster the data. Furthermore, it has no mechanism to avoid Feature Randomness after discarding the decoder. Hence, the mass of random labels can lead to the generation of random discriminative features. Besides, the contribution of each center to the loss function is normalized to block large clusters from altering the embedded representations. Therefore, DEC is more suitable for balanced datasets.

Improved Deep Embedded Clustering (IDEC) \cite{paper28} is quite similar to DEC. The major contribution of IDEC is balancing the clustering cost with a reconstruction one after the pretraining phase. According to Xifeng et al. \cite{paper28}, eliminating the reconstruction during the clustering phase hampers the network from preserving the local structures. However, we argue that maintaining the reconstruction loss end-to-end is not as beneficial due to the inherent conflict between clustering and reconstruction. More precisely, embedded clustering does not take into consideration non-discriminative patterns that do not contribute to data categorization (e.g. between-class similarities), whereas reconstruction is mainly concerned with preserving all the information whether discriminative or not. Preserving non-discriminative features end-to-end implies a reduction in the overall discriminative ability.

Deep Clustering Network (DCN) \cite{paper29} is also an autoencoder clustering approach. Similar to DEC and IDEC, DCN has a joint optimization process. First, the autoencoder is pretrained to reduce the dataset dimensionality based on a reconstruction loss function. The ultimate goal of this approach is to get a K-Means-friendly representation at the end of the training process. To this end, a K-Means loss is applied along with the vanilla reconstruction. This method requires hard clustering assignments (as opposed to soft clustering assignments based on probabilities). That would induce a discrete optimization process, which is incongruous with the differential aspect of gradient descent. Similar to IDEC, DCN does not have any mechanism to mitigate the clustering-reconstruction trade-off.

Joint Unsupervised LEarning (JULE) \cite{paper22} is a deep recurrent framework for jointly learning unsupervised representations using CNN and clustering images using agglomerative clustering. The main contribution of JULE is the recurrent process that allows to merge clusters in multiple time steps. However, training a recurrent neural network, where the number of time-steps is equal to the number of data points, is not computationally efficient.

Deep Embedded Regularized Clustering (DEPICT) \cite{paper36} leverages a convolutional autoencoder for learning embedded representations and their clustering assignments. Similar to DEC, DEPICT has a relative cross-entropy (KL divergence) objective function. In addition, the loss function of DEPICT has a regularization term, which allows to explicitly impose the size of each cluster based on some prior knowledge. This avoids situations, where most of the data points are assigned to a few clusters. However, this prior knowledge (size of clusters) assumed by DEPICT is impractical for pure unsupervised problems. Added to that, DEPICT overlooked the clustering-reconstruction trade-off.

Variational Deep Embedding (VaDE) \cite{paper35} is a generative deep clustering method, inspired by Variational Autoencoder (VAE). It allows to couple clustering with data generation. The VaDE framework involves encoding the initial data distribution as a \textit{Gaussian Mixture distribution} in the embedded space. In order to allow backpropagation, the \textit{Gaussian Mixture distribution} is sampled using the reparameterization trick. VaDE relies on variational inference. Thus, the information loss induced by the mean-field approximation can harm the quality of the latent space, which in turn may deteriorate the clustering performance.

Adversarial Deep Embedded Clustering (ADEC) \cite{paper98} is an autoencoder-based clustering model. It is the first work to characterize and address the trade-off between Feature Randomness and Feature Drift. Similar to DEC, ADEC minimizes the Kullback–Leibler (KL) divergence to an auxiliary target distribution. Additionally, the clustering objective of ADEC is regularized by an adversarially constrained reconstruction. Based on adversarial training, the strong competition between the two objective functions is alleviated. However, adversarial training is still a challenging task due to its lack of stability. This instability can be explained by the following reasons: (1) Mode collapse \cite{paper2}: this happens when the generated data distribution only captures a local region from the real data support, which will induce restricted varieties of samples, (2) Failure to converge \cite{paper99}, and (3) Memorization: this problem takes place when the model can not synthesize new data points different from the ones fed to the network during the training process. By optimizing a dynamic objective function, our empirical results show that it is possible to find a better trade-off between Feature Randomness and Feature Drift compared to ADEC, without using adversarial training, and without adding an additional network (i.e., discriminator).

\section{Motivations}

In this section, we review the trade-off between Feature Randomness and Feature Drift \cite{paper98}, which serves as the motivation behind our proposed method. As mentioned before, the main challenge in unsupervised learning is that there is no obvious straightforward objective function that can learn high-level similarities without feeding semantic labels. Existing unsupervised representation learning strategies stem from two concepts: self-supervision and pseudo-supervision.

Pseudo-supervision consists of training a model using pseudo-labels \cite{paper87} instead of the ground truth ones. Unlike self-supervision, a pseudo-supervised objective function targets \textit{explicitly} the semantic categories without using any proxy. Similar to self-supervision, the computed labels are automatically extracted without any human intervention. In practice, the pseudo-labels are predicted based on uncertain assumptions and hypothetical similarities. Therefore, some of them mismatch the real ones. In this work, a pseudo-supervised loss is denoted by $L_{P}$.

Self-supervision targets learning general-purpose features by solving a \textit{pretext} task. To this end, the training data is automatically labeled by finding and exploiting simple intrinsic information (e.g., color, position, context). In practice, self-supervised approaches have competitive results on multiple benchmarks \cite{paper60, paper93}. The success of self-supervision is linked to the selected \textit{pretext}. In order to be solved, a good \textit{pretext} problem is designed in a way that requires learning high-level features. Multiple \textit{pretext} objectives were proposed in the literature. The simplest one is the reconstruction \textit{pretext}. Also, popular \textit{pretext} tasks include the adversarial objective \cite{paper1} that relies on distinguishing fake and real samples and the denoising objective \cite{paper58}. In computer vision, we can find the following pretexts: predicting the colorization \cite{paper63}, predicting unpainted patches \cite{paper62}, predicting the spatial relationship of different patches \cite{paper59}. An interesting overview of existing self-supervised methods can be found in \cite{paper94}. In this work, a self-supervised loss is denoted by $L_{S}$.

\subsection{Feature Randomness}
Existing clustering models enable classifying the data without using labels. However, supervised learning approaches yield much better results than the clustering ones. This disparity can be explained by the few assumptions made by the existing clustering methods. These assumptions are rarely congruent with real data complexity. In the absence of concrete labels, it is possible to train a neural network using predicted labels that were extracted based on hypothetical assumptions. Obviously, these constructed labels are less effective than true labels because some of them mismatch the real categorization. Zhang et al. \cite{paper66} trained several standard deep neural networks architectures with different levels of random labels. The most important observation of this study is that all the trained models achieve a training error equal to zero regardless of the level of randomness. Surprisingly, increasing the portion of random labels does not cause any significant training time overhead. These findings confirm that a deep neural network can perfectly fit random labels when the number of its parameters exceeds the number of data points. An implication of this conclusion is that the capacity of a neural network is adequate to memorize the whole dataset. Furthermore, the correlation between the training samples and their corresponding labels has little to do with the training feasibility.

Feature Randomness occurs when a neural network is trained using pseudo-supervised labels.
At every iteration, Feature Randomness is characterized by the deviation of the gradient of the real supervised objective function w.r.t the network parameters $w$ after introducing pseudo-labels. Mathematically, Feature Randomness can be expressed as the cosine of the angle between the gradient of the real supervised objective function and the gradient of the unsupervised objective function. This characterization $\Delta_{FR}$ is described by Equation \ref{eq:delta_FR}, where $x$ is the input data, L is the loss function, $y_{true}$ is the vector of true labels and $y_{pseudo}$ is the vector of pseudo labels.

\begin{equation} \label{eq:delta_FR}
    \Delta_{FR}= cos(\widehat{\frac{\partial L(x, y_{true}, w)}{\partial w}, \frac{\partial L(x, y_{pseudo}, w)}{\partial w}}).
\end{equation}

Training with pseudo-labels pushes the neural network to learn features that emphasize similarities between data points from different clusters. It also enforces learning to exclude points from their natural clusters and map them to irrelevant clusters. These implications give birth to features that contradict the data semanticity. Hence, they are considered to be random features. 

We can compare $ACC$, which is the standard metric for evaluating deep clustering models, with $\Delta_{FR}$. While $ACC$ computes the similarity between the two \textit{scalar} values $y_{true}$ and $y_{pseudo}$,  $\Delta_{FR}$ computes the similarity between the gradient \textit{vectors} inherent to $y_{true}$ and $y_{pseudo}$. Obviously, the gradient vector provides a better insight of the topological characteristics of the latent space. Hence, $\Delta_{FR}$ can be used to identify problems related to deep clustering (e.g., Feature Randomness), whereas $ACC$ falls short of this capacity. Our experimental visualisations using both metrics strongly supports this intuition.

As mentioned before, a self-supervised objective function targets learning general-purpose features by solving a \textit{pretext} task. Although the learned features are less discriminative than the ones learned based on categorical labels (i.e., labels that indicate the high semantic categories), for a self-supervised objective function, there are no pseudo-labels. Therefore, $y_{true} = y_{pseudo}$. Hence,  $\Delta_{FR}$ is always equal to 1 and there is no possibility of Feature Randomness to take place. Thus, it is possible to mitigate Feature Randomness by combining a pseudo-supervised objective function with a self-supervised one. Furthermore, the self-supervised objective function can be used to incorporate relevant prior knowledge (e.g., invariance to small translations and rotations for images datasets).



\subsection{Feature Drift}

State-of-the-art autoencoder-based clustering approaches rely on jointly performing representation learning and clustering based on a linear combination of two objective functions. The joint optimization process can be described as follows:

\begin{equation}
    L = L_{1} + \gamma \: L_{2}.
\end{equation}

\begin{figure}[ht]
\vskip 0.2in
\begin{center}
\centerline{\includegraphics[width=250pt, height=120pt]{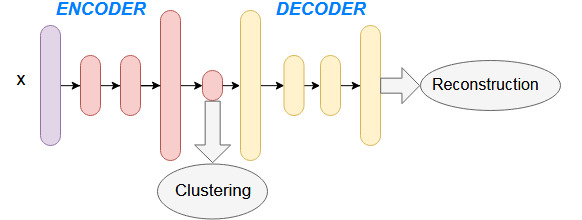}}
\caption{The common network structure of autoencoder-based clustering algorithms.}
\label{fig:architecture_sota}
\end{center}
\vskip -0.2in
\end{figure}

Where $L_{1}$ is the reconstruction loss function and $L_{2}$ is the embedded clustering loss. $\gamma$ is a hyper-parameter, which is used to balance the two costs. It has been shown empirically \cite{paper28} that it is better to keep $\gamma$ small in order to avoid having the clustering cost corrupt the latent space. The common network architecture of the autoencoder-based clustering approaches is illustrated in Figure \ref{fig:architecture_sota}.

In the context of multi-objective optimization, decisions are made by taking into consideration trade-offs between conflicting objectives. Objective functions are said to be conflicting if there is no solution simultaneously optimizing every single objective function. Optimizing an objective function can degrade another one in value when they are in conflict. In such a case, it is possible to get an infinite number of Pareto optimal solutions. Therefore, additional subjective information is required to select a single solution. Without acquiring this information, all solutions are equally valid.

Feature Drift takes place when a neural network is trained to minimize a combination of two \textit{strongly} conflicting loss functions. In the specific context of deep clustering, the two cost functions are the pseudo-supervised loss and the self-supervised one. 

At every training iteration, Feature Drift is characterized by the degree of competition between the gradient of the self-supervised objective function and the gradient of the pseudo-supervised objective function w.r.t the network parameters $w$. Mathematically, Feature Drift can be expressed as the cosine of the angle between theses two gradient vectors. This characterization $\Delta_{FD}$ is described by Equation \ref{eq:delta_FD}, where $y_{pretext}$ is the self-acquired labels for the \textit{pretext} task and $y_{pseudo}$ is the vector of pseudo labels. Training a neural network with two strongly conflicting objective functions have its drawbacks. In fact, the features learned by optimizing the main objective function can be easily drifted by updating the learning weights with respect to the secondary objective function. This can make the global optimization process fail. The illustrative example from \cite{paper98} offers a considerable insight into this problem.

\begin{equation} \label{eq:delta_FD}
    \Delta_{FD}= cos(\widehat{\frac{\partial L_{P}(x, y_{pseudo}, w)}{\partial w}, \frac{\partial L_{S}(x, y_{pretext}, w)}{\partial w}}).
\end{equation}

In a typical autoencoder-based clustering model, rising the balancing coefficient significantly makes the pseudo-supervised objective function dominate the optimization process. In such a case, Feature Randomness increases. However, decreasing the balancing coefficient too much makes the self-supervised gradient dominate the competition. This means that the discriminative representations learned by pseudo-supervision can be easily drifted by the self-supervised objective function. Hence, we try to balance the trade-off between Feature Randomness and Feature Drift in our proposed method.

\section{Proposed Dynamic Autoencoder Model}
\label{Sec:methodoloy}
In this section, we present our double-stage deep clustering model. In section \ref{Sec:methodoloy}.1, we describe the pretraining phase. Pretraining allows us to learn general-purpose features to initialize the neural network. As stated before, pseudo-labels are unreliable when extracted from random latent representations. It is reasonable to predict that completely random pseudo-labels are not a good starting supervisory signal. It leads to excessive Feature Randomness, which in turn misleads the whole learning process. Hence, before starting to train using a pseudo-supervised objective function, the neural network can be pretrained to learn valuable information relying on some \textit{pretext} task. After that, learning discriminative representations becomes more suitable by introducing effective pseudo-supervision. Based on that, any self-supervised loss is a candidate for the pretraining phase. In our case, we opted for reconstruction regularized by adversarially constrained interpolation and data augmentation for their competitive results compared to state-of-the-art self-supervised approaches. In Section \ref{Sec:methodoloy}.2, we describe our dynamic loss function, which is the main contribution of this work. Our dynamic learning mechanism offers a better trade-off between Feature Randomness and Feature Drift by gradually and smoothly eliminating the reconstruction objective function in favor of a construction one. 

Before starting our methodology description, we establish some useful notations.
Consider the problem of clustering a dataset  $ X = \left \{ x_{i} \in \mathbb{R}^{d} \right \}_{i=1}^{N}$ of $N$ data points in $\mathbb{R}^{d}$. This dataset can be grouped into $K$ clusters $\left \{ C_{k} \right \}_{k=1}^{K}$. $f_{\theta_{e}}:X\rightarrow Z$ and  $g_{\theta_{d}}:Z\rightarrow X$ stand for the encoder and decoder mappings, respectively and $\theta_{e}$ and $\theta_{d}$ represent their learnable parameters, respectively. Let $z_{i} = f_{\theta_{e}}(x_{i}) \in \mathbb{R}^{p} $ be the latent representation of the data point $x_{i}$ and $\hat{x_{i}} = g_{\theta_{d}}(z_{i}) \in \mathbb{R}^{d} $ be the reconstructed representation of $x_{i}$. The number of cluster $K$ is provided as prior knowledge and $\Gamma  = \left \{ \mu_{j} \in  \mathbb{R}^{p} \right \}_{j=1}^{K}$ is the set of latent centroids. In order to assess the similarity between the latent centroids $\mu_{j}$ and the embedded data points $z_{i}$, the Student’s t-distribution is selected as a kernel. The same kernel was selected for other deep clustering approaches, such as, DEC \cite{paper27} and IDEC \cite{paper28}. Let $q_{ij}$ denotes the soft clustering assignments based on the Student’s t-distribution kernel as shown in Equation \ref{eq:q_ij}. It accounts for the probability of assigning an embedded point $z_{i}$ to the centroid $\mu_{j}$. $\alpha$ is a coefficient related to the Student’s t-distribution. Based on the matrix $Q = (q_{ij})_{i = 1...N, j = 1...K}$, we define the function $\sigma : X \rightarrow  Z$, where $\sigma(x_{i})$ outputs the centroid with the highest clustering assignment. Hence, $\sigma(x_{i}) = \mu_{argmax_{j}\left \{ q_{ij} \right \}}$.

\begin{equation} \label{eq:q_ij}
  \begin{aligned}
    q_{ij} = \frac{(1 + \frac{\left \| z_{i} - \mu_{j} \right \|^{2}}{\alpha })^{-\frac{\alpha+1}{2}}}{\sum_{j'}(1 + \frac{\left \| z_{i} - \mu_{j'} \right \|^{2}}{\alpha })^{-\frac{\alpha+1}{2}}}
  \end{aligned},
\end{equation}

For each data point $x_{i}$, we define $h_{ij'}$ as the $j'^{th}$ maximal value of $q_{ij}$. For example, $h_{i1}$ is the maximal value of $q_{ij}$ and $h_{i2}$ is the second maximal value of $q_{ij}$. Mathematically, $h_{ij'}$ can be formalized as shown in Equation \ref{eq:h}.

\begin{equation}\label{eq:h}
    \begin{aligned}
    h_{ij'} =\left\{
                \begin{array}{ll}
                  max_{j}\left \{ q_{ij} \right \}, \; if  \;  j'=1,
                  \\
                  max_{j}\left \{ q_{ij} \; | \; q_{ij}<h_{i(j'-1)} \right \},  \; otherwise.
                \end{array}
              \right. 
    \end{aligned}
\end{equation}

The ultimate goal of our methodology is to find $\theta_{e}$ that allows the $f$ transformation to project the data in a clustering-friendly embedded space, according to the selected kernel.

\subsection{Pretraining}

Similar to all autoencoder-based clustering methods, our dynamic autoencoder model has a pretraining phase. Therefore, the encoder and decoder are trained to optimize a \textit{pretext} objective. Then, the learning weights can be fine-tuned according to the main objective function. Previous deep clustering models, such as, \cite{paper27} and \cite{paper28}, leverage a stacked denoising auto-encoding strategy for learning general-purpose representations. It is important to highlight that pretraining with a self-supervised objective function allows capturing embedded features related to the data distribution. It then allows avoiding extracting pseudo-labels from completely random latent representations. In other words, pretraining is a keystone in reducing Feature Randomness.

To better understand the purpose of a double-stage framework, we provide an analogy with signal processing. Extracting discriminative representations can be considered similar to a filtering operation. In signal processing, a filter allows to destroy irrelevant components and preserve pertinent patterns. To this end, the signal should be expressed in the frequency domain before filtering the undesired information. The Fourier transform is a bijective function. It allows mapping the initial signal to the frequency domain, where spectral components are expressed. Similar to the Fourier transform, the pretraining phase relies on a generative transformation, where the latent representations can be inverted to their initial form. Then, based on the obtained representations, discriminative patterns can be learned by bringing the embedded data points closer to their associated centroids. This leads to destroying unwanted information, namely, within-cluster similarities. This operation is similar to filtering in signal processing. The simple analogy described above is summarized in Table \ref{table:1}.

\begin{table}[htb]
\caption {Analogy between autoencoder-based clustering and signal processing.}
\label{table:1}
\begin{center}
\begin{small}
\begin{tabular}{ |>{\centering\arraybackslash}m{2.1cm}|>{\centering\arraybackslash}m{3.5cm}|>{\centering\arraybackslash}m{3cm}|>{\centering\arraybackslash}m{3cm}|}
        \hline
        \textbf{Signal \newline processing}  & \textbf{Autoencoder clustering}  & \textbf{Transformation} &  \textbf{Strategy}  \\ \hline
        Fourier \newline transform & Reconstruction & Generative \newline transformation & Target patterns are more pronounced in the new space. 
        \\ \hline
        Filtering & Embedded clustering & Discriminative \newline transformation & Non discriminative patterns are destroyed.
        \\ \hline
    \end{tabular}
\end{small}
\end{center}
\end{table}

Unlike previous autoencoder-based clustering methods, our pretraining phase consists of optimizing a vanilla reconstruction objective function regularized by data augmentation (e.g., small random shifting and small random rotation) \cite{paper34} and an adversarially constrained interpolation \cite{paper5}. Both techniques showed competitive results in learning suitable unsupervised representations \cite{paper34, paper5}. 

The ACAI (i.e., Adversarially Constrained Autoencoder Interpolation) framework consists of a game competition between two adversarial networks, namely, an autoencoder and a critic. The purpose of this framework is to make the autoencoder generate samples that have a mixture of semantic characteristics from the inputs. For every training iteration, two coefficients $\alpha$ and $\gamma$ are randomly sampled from the interval $[0,1]$. Furthermore, a couple of data points $x_{1}$ and $x_{2}$ needs to be sampled randomly to compute $x_{\alpha} = g_{\theta_{d}}(\alpha f_{\theta_{e}}(x_{1}) + (1-\alpha) f_{\theta_{e}}(x_{2}))$. As the mathematical equation indicates, $x_{\alpha}$ represents the reconstructed data point whose latent representation is linearly interpolated from the embedded representation of $x_{1}$ and $x_{2}$. $c$ denotes the critic network and $\theta_{c}$ denotes the learnable parameters of this network. t stands for the index of the iteration. While the critic is trained to regress the interpolation coefficient $\alpha$ from $\hat{x}_{\alpha}$ in Equation \ref{eq:L_{c}}, the main network is trained to reconstruct the input data and to fool the critic into considering the interpolated points to be realistic in Equation \ref{eq:L_{f,g}}. The second term in Equation \ref{eq:L_{c}} enforces the critic to output 0 for non-interpolated inputs. This regularization technique makes the interpolants look realistic, which leads to a smooth latent space as shown by \cite{paper5}. To keep the notation simple, we consider that $x$ represents the data samples after performing some random transformations (translation and rotation). The whole pretraining framework is shown in Figure \ref{fig:pretraining}.

\begin{equation}\label{eq:L_{f,g}}
    L_{f,g}(\theta_{e}(t), \theta_{d}(t)) =  ||x - \hat{x}||_{2}^{2} + \lambda \: ||c(\hat{x}_{\alpha})||_{2}^{2},
\end{equation}

\begin{equation}\label{eq:L_{c}}
    L_{c}(\theta_{c}(t)) = ||c(\hat{x}_{\alpha}) -\alpha||_{2}^{2} + ||c(\gamma x + (1-\gamma)\hat{x})||_{2}^{2}.
\end{equation}

\textbf{\begin{figure}[ht]
\vskip 0.2in
\begin{center}
\centerline{\includegraphics[width=\columnwidth, height=240pt]{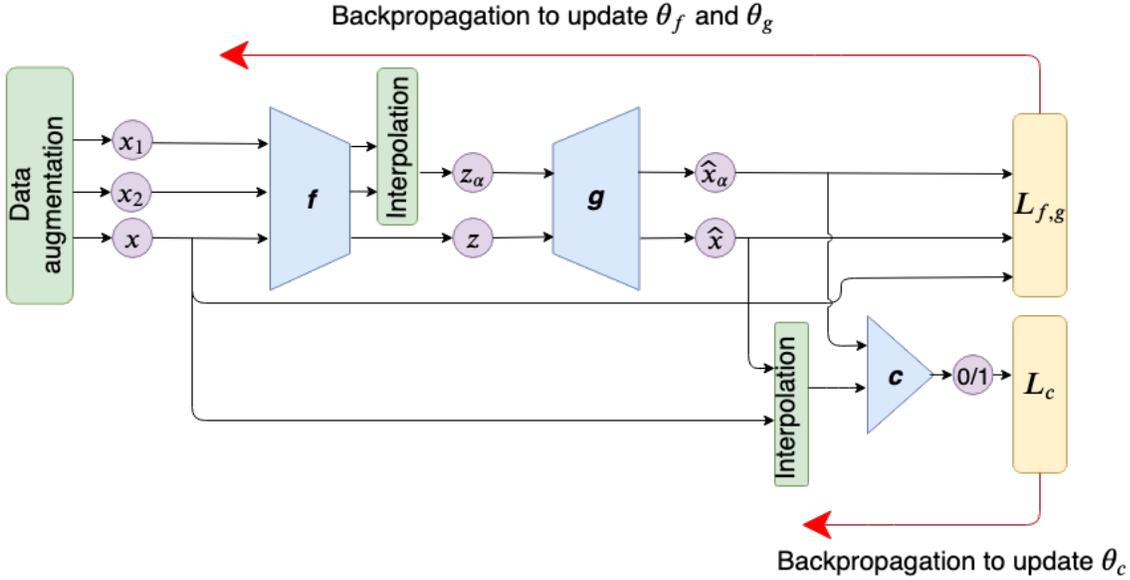}}
\caption{The pretraining phase of DynAE.}
\label{fig:pretraining}
\end{center}
\vskip -0.2in
\end{figure}}

\subsection{Clustering phase}
\begin{figure}[ht]
\vskip 0.2in
\begin{center}
\centerline{\includegraphics[width=300pt, height=120pt]{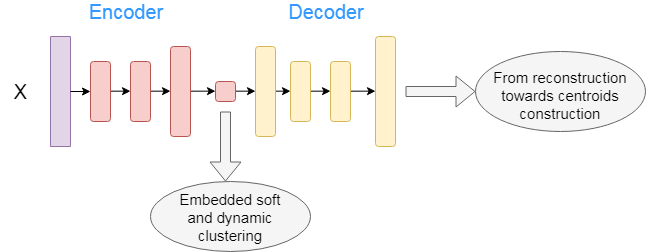}}
\caption{The general clustering architecture of DynAE.}
\label{fig:clustering_diagram}
\end{center}
\vskip -0.2in
\end{figure}

After pretraining, we finetune the autoencoder weights by optimizing a dynamic loss. To this end, we use K-Means to initialize the embedded clustering centers. Apart from K-Means, there are multiple popular strategies for computing centroids' coordinates, such as k-medoids \cite{paper95} and clarans \cite{paper96}. In this work, for the sake of simplicity and since the goal is to show empirically that DynAE is a promising deep clustering strategy, K-Means is selected for initializing and updating the centroids. 
Similar to state-of-the-art autoencoder-based clustering models, our cost function has two parts. The first one makes up the reconstruction cost and the second one is the clustering objective function. Figure \ref{fig:clustering_diagram} illustrates the general clustering architecture of DynAE.
 
\subsubsection{From reconstruction towards centroids construction}

Our idea is to make the autoencoder output the centroid image of every sample instead of generating reconstructed images. To this end, the decoder can be used to generate the centers' images from the embedded centroids $\mu_{j}$. Then, we can feed them to the network as a supervisory signal. However, abrupt elimination of the self-supervised loss (i.e., reconstruction) can lead to Feature Randomness, whereas, preserving it during the whole training process causes Feature Drift. We propose to gradually and smoothly eliminate the reconstruction loss in favor of centroids construction. 

After the pretraining phase, some embedded data points $z_{i}$ have unreliable clustering assignments $q_{ij}$. More explicitly, the probabilities of their highest assignments are very close to each other. Generally, these samples are located near the cluster boundaries. Therefore, it is hard to predict their corresponding centers confidently. We consider them as conflicted data points. At every training iteration, we characterize an unconflicted data point by a condition that controls the level of acceptable uncertainty. Mathematically, a conflicted data point belongs to the set $\overline{S}$ defined by Equation \ref{eq:S_bar}. The hyperparameters $\beta_{1}$ and $\beta_{1}$ are controlling thresholds. They lie in the interval $[0, 1]$. $\beta_{1}$ stands for the minimal confidence threshold under which a data point is considered to be conflicted. $\beta_{2}$ is the minimal difference between the highest and the second-highest assignment probabilities ($h_{i1}$ and $h_{i2}$, respectively, as defined by Equation \ref{eq:h}). Based on our definition, a conflicted sample must have its highest assignment less than $\beta_{1}$ or the difference between its highest and second-highest assignments must be less than the threshold $\beta_{2}$. 

\begin{equation}\label{eq:S_bar}
    \overline{S}=\left \{ x_{i} \in X | \; h_{i1} <  \beta_{1} \; or \;  (h_{i1}-h_{i2}) <  \beta_{2}  \right \}. 
\end{equation}

The first part of our dynamic loss function is described by Equation \ref{eq:L_{1}}. Reliable samples are selected for centroid construction. As for the conflicted data points, the reconstruction cost is applied until the model gains more confidence in their corresponding centers. So depending on the data sample, there are two possible training schemes: reconstruction or centroid construction as stated by Equation \ref{eq:L_{1}}.

\begin{equation} \label{eq:L_{1}}
    \begin{aligned}
    L_{1}(\theta_{e}(t), \theta_{d}(t)) = \sum_{i=1}^{N} \left\{
                \begin{array}{ll}
                  ||x_{i} - \hat{x_{i}}||_{2}^{2} \;\;\; if \; x_{i} \; \in \; \overline{S},
                  \\
                  ||g(\sigma(x_{i})) - \hat{x_{i}}||_{2}^{2} \;\;\;  otherwise.
                \end{array}
              \right. 
    \end{aligned}
\end{equation}

The centroids images, which are used as a supervisory signal, are generated using the decoder and do not represent real data points. In order to avoid blurry images, the first nearest neighbor (1-NN) of each embedded center $\mu_{j}$ is selected as a substitute to the decoder's generated images. It is possible that pretraining with ACAI contributes to generating stable centroids by introducing continuity in the latent space \cite{paper5}.

Generally, hyperparameters are known to be dataset-specific. Besides, for real unsupervised applications, supervised cross-validation is impractical. Therefore, it is vital to make our method less sensitive to the choice of unpredictable hyperparameters. In our case, $\beta_{1}$ and $\beta_{2}$ are computed based on mathematical formulations. As shown by Equation \ref{eq:beta_1_and_beta_2}, they depend on $\kappa$, which is the confidence threshold.  

\begin{equation}\label{eq:beta_1_and_beta_2}
    \beta_{1} = \frac{\kappa}{K} \;\; and \;\; \beta_{2} = \frac{\beta_{1}}{2},\;\;\; \kappa \in [|1,K|].
\end{equation}

It is worth to note that $\kappa$ is initialized in a way to ensure that $\beta_{1}$ and $\beta_{2}$ have high values. Thus, the controlling condition for unconflicted data becomes very difficult to satisfy. At the initial stage, most of the training set is devoted to being reconstructed and the rest is used for performing centroids construction. Starting with higher values would allow for more reconstruction and fewer centroids construction, which would only slow down convergence. 

We update $\kappa$ in a way that captures the dynamics of the number of conflicted data points. As shown by Equation \ref{eq:tau}, we can describe the dynamics of the loss function by $\tau$, which specifies the amount of reconstruction in Equation \ref{eq:L_{1}}. $\tau$ is not a tunable hyperparameter, but instead just an indicator of the training progress.

\begin{equation}\label{eq:tau}
    \tau = \frac{nb \: of \: reconstructed \: samples}{N} = \frac{\mid\overline{S}\mid}{N}.
\end{equation}

During the training process, the number of conflicted points is supposed to decrease gradually. The dynamic loss function reaches stability when the number of conflicted data points does not decrease any more. At this level, $\tau$ remains constant.

The local convergence is characterized by the stability of the loss function. In order to escape local convergence, there are two options. The first option is to decrease the hyperparameters $\beta_{1}$ and $\beta_{2}$ and the second one consists of updating the centroids. In this work, we opt for both solutions to avoid local convergence. The hyperparameters $\beta_{1}$ and $\beta_{2}$ are dropped by $\frac{\triangledown\kappa}{K}$, where $\triangledown\kappa$ is the dropping rate of $\kappa$.

\begin{figure}[ht]
\vskip 0.2in
\begin{center}
\centerline{\includegraphics[width=\columnwidth]{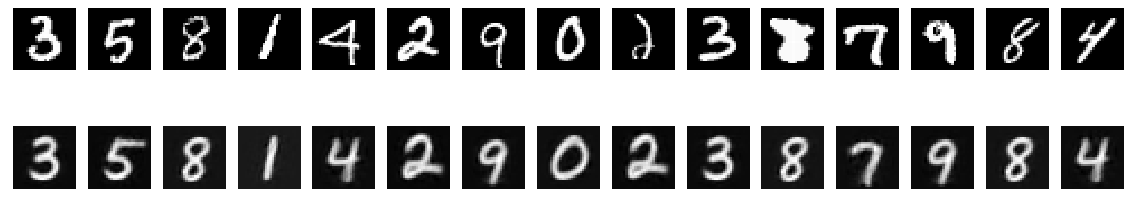}}
\caption{The centroids construction from DynAE at the end of the training process. Top row: MNIST input; bottom row: Output images from DynAE.}
\label{fig:new_output}
\end{center}
\vskip -0.2in
\end{figure}

We train our model until no reconstruction remains. At the end of the training process, each sample is mapped to its corresponding centroid. The output images of the network are smoother and easier to recognize as illustrated in Figure \ref{fig:new_output}.

\subsubsection{Embedded clustering}

Embedded data points of each cluster are generally spread near non-linear manifolds. However, typical centroid-based clustering algorithms (e.g., K-Means) are only suitable for blob-like distributions. Minimizing the $L_{1}$ loss function does not make the embedded space clustering-friendly. To deal with this problem, an embedded clustering loss function $L_{2}$ is proposed to make the latent representations liable to centroid-based clustering. To this end, we propose an embedded clustering cost function that penalizes the distance between the data points and their associated centroids. Hence, the encoder learns to project the samples closer to their corresponding clustering representatives. This strategy was first introduced to train DCN \cite{paper29}. However, the aforementioned model has hard clustering assignments and does not consider any adaptive learning dynamics. All the embedded samples are pulled towards their corresponding centroids without considering the uncertainty of the clustering assignments. 

Our proposed embedded clustering objective function is described by Equations \ref{eq:L_{2}} and \ref{eq:S}.  Compared to DCN, we use soft clustering assignments based on the Student’s t-distribution. Moreover, our cost function is more suitable for handling the conflicted data points issue. Similar to the previous loss function in Equation \ref{eq:L_{1}}, the learning dynamics are controlled by the same confidence hyperparameters $\beta_{1}$ and $\beta_{2}$. The set of reliable samples (i.e., unconflicted) is defined by Equation \ref{eq:S}. The number of points belonging to this set increases automatically from one iteration to another. This happens even if we keep  $\beta_{1}$ and $\beta_{2}$ constant during the whole training process. This result is confirmed by the experiments presented in the following section. In the beginning, simple patterns of each cluster are learned by attracting the most reliable examples to their corresponding centroids. Then, based on the knowledge acquired from clustering and attracting the most reliable samples, more difficult examples become reliable to be clustered. Unlike $L_{1}$, conflicted data points are not exploited in $L_{2}$ until they become unconflicted. 

\begin{equation} \label{eq:L_{2}}
    L_{2}(\theta_{e}(t), \theta_{d}(t)) = \sum_{x_{i}\in S} \left \| f_{\theta_{e}}(x_{i})- \sigma(x_{i})) \right \|_{2}^{2},
\end{equation}

\begin{equation}\label{eq:S}
    S=\left \{ x_{i} \in X | \; h_{i1} \geq  \beta_{1} \; and \;  (h_{i1}-h_{i2}) \geq   \beta_{2}  \right \}. 
\end{equation}

\subsubsection{Joint clustering and centroids construction}

The Complete dynamic objective function is defined in Equation \ref{eq:L}. Similar to the pretraining phase, the loss function L is regularized by data augmentation. Unlike related works \cite{paper27, paper28, paper29, paper36}, the balancing hyperparameter is unrequired in our case. DynAE aims to reach a better trade-off between Feature Randomness and Feature Drift using the learning dynamics (without any balancing hyperparameter). The whole clustering framework is shown in Figure \ref{fig:clustering}.

\begin{equation}\label{eq:L}
    L(\theta_{e}(t), \theta_{d}(t)) = L_{1}(\theta_{e}(t), \theta_{d}(t)) + L_{2}(\theta_{e}(t), \theta_{d}(t)).
\end{equation}

\textbf{\begin{figure}[ht]
\vskip 0.2in
\begin{center}
\centerline{\includegraphics[width=\columnwidth, height=200pt]{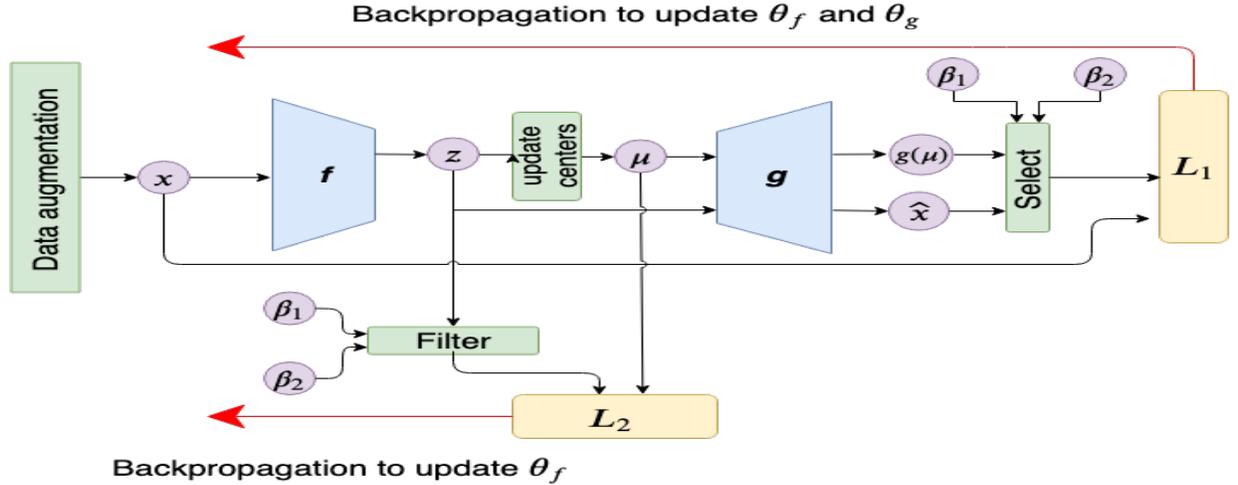}}
\caption{The clustering phase of DynAE.}
\label{fig:clustering}
\end{center}
\vskip -0.2in
\end{figure}}

\subsection{Optimization}

The loss function \ref{eq:L} is optimized using mini-batch gradient descent and backpropagation to solve for the autoencoder's weights. We run the optimization process for $MaxIter$ batch iterations or until $\tau$ becomes lower than a specific amount $tol$ of the total training set. 

During DynAE training, our objective function moves smoothly and gradually from pure self-supervision towards pure pseudo-supervision. In the beginning, the model is pretrained with a purely self-supervised objective function. In the end, the final objective function is almost purely pseudo-supervised ($\tau$ becomes very low). This transition from self-supervision towards pseudo-supervision includes transient objectives, where we combine both strategies. As the training progresses, pseudo-supervision increases and self-supervision decreases. The intuition behind our idea (described in detail earlier) can be summarized as: a) Pseudo-supervision is not reliable at the beginning due to a large number of fake labels. It leads to Feature Randomness. b) Self-supervision is great for learning general-purpose features, but it does not reflect our ultimate clustering objective. c) Combining pseudo-supervision and self-supervision can cause Feature Drift. d) Pseudo-supervision is very effective when we have reliable pseudo-labels. Our dynamic training strategy is illustrated in Figure \ref{fig:dynamic_training}.

\textbf{\begin{figure}[ht]
\vskip 0.2in
\begin{center}
\centerline{\includegraphics[width=\columnwidth, height=110pt]{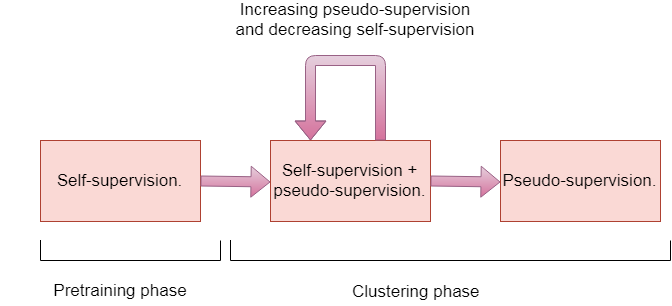}}
\caption{An illustration of our complete training strategy.}
\label{fig:dynamic_training}
\end{center}
\vskip -0.2in
\end{figure}}

Our method is illustrated in Algorithm \ref{algorithm1}. Without considering the pretraining phase, the computational complexity of DynAE is $\mathcal{O}(mLD^{3})$, where $m$ is the number of training iterations, $L$ is the number of layers and $D$ is the maximal number of neurons in the hidden layers. It is worth to mention that DEC, IDEC, DCN and DynAE, all have the same computational complexity. Therefore, for the same network architecture, the same batch size, the same optimizer and the same pretraining phase, the execution time of these four approaches can be strictly compared based on the number of iterations required for convergence.

\begin{algorithm*}
    \caption{Dynamic autoencoder.}
    \label{algorithm1}
\begin{algorithmic}
    \STATE {\bfseries Input:} Input data: $X$, Number of clusters: $K$, Maximaum iterations: $MaxIter$, Learning rate: $\vartheta $, Convergence threshold: $tol$, Confidence threshold: $\kappa$.
    \STATE {\bfseries Output:} Autoencoder weights: $W$
    \STATE Initialize autoencoder weights using our pretraining strategy.
    \STATE Initialize embedded centroids $\Gamma$ using kmeans.
    \STATE $\beta_{1} \gets \frac{\kappa}{K}$ 
    \STATE $\beta_{2} \gets \frac{\beta_{1}}{2}$
    \STATE nbConfPrev $ \gets |X|$ $\: \: \triangleright$ nbConfPrev: previous number of conflicted points.
    \FOR{$i=0$ {\bfseries to} $MaxIter$}
        \IF{$i > 0$}
            \STATE nbConfPrev $ \gets $ nbConf $\: \: \:  \triangleright$ nbConf: number of conflicted points.
        \ENDIF
        \STATE nbConf $ \gets$ numberConflictedPoints$(X, \Gamma, \beta_{1}, \beta_{2})$
        \IF{nbConf $\geq$ nbConfPrev}
            \STATE $\Gamma \gets$ updateCentroids$(X, K)$
            \STATE $\beta_{1} \gets \beta_{1} -  \frac{1}{K} \triangledown\kappa$
            \STATE $\beta_{2} \gets \beta_{2} - \frac{1}{K} \triangledown\kappa$
            \STATE $\kappa \gets \triangledown\kappa$
        \ENDIF
        \IF{$\frac{nbConf}{|X|} < tol$}
            \STATE End Training
        \ENDIF
        \STATE Compute $L$ according to Equation \ref{eq:L} 
        \STATE $W \gets W - \vartheta \frac{\partial L}{\partial W}$
    \ENDFOR
\end{algorithmic}
\end{algorithm*}

\section{Experiments}

\label{Sec:experiments}
As explained in the previous section, our methodology consists of optimizing a dynamic objective function to gradually and smoothly reach a better trade-off between Feature Randomness and Feature Drift. In order to validate the suitability of smooth dynamics in solving the stated problem, we conduct extensive experiments.
In section \ref{Sec:experiments}.1, we present the experimental settings. In section \ref{Sec:experiments}.2, we present our results, which strongly back-up our intuitions. 

\subsection{Experimental Settings} 
All experiments were carried out on Google Colaboratory platform \footnote{\textit{\url{https://colab.research.google.com}}}. The code of our model DynAE is published on Github \footnote{ \textit{\href{https://github.com/nairouz/DynAE}{https://github.com/nairouz/DynAE}}}.

\subsubsection{Datasets and baselines}

We compare DynAE with several clustering algorithms on four famous datasets: MNIST-full \cite{paper38}, MNIST-test, USPS \cite{paper39}, and Fashion-MNIST \cite{paper40}. All samples were normalized and flattened before feeding them to our fully-connected network. For convolutional deep clustering models, the input data is not flattened.

\begin{itemize}
    \item MNIST-full \cite{paper38}: A 10 classes dataset of $70000$ samples. Each sample is a $28\times28$ grayscale image of a handwritten digit. 
    \item MNIST-test: A test subset of the MNIST-full dataset with $10000$ data samples.  
    \item USPS \cite{paper39}: A 10 classes dataset of $9298$ samples. Each sample is a $16\times16$ grayscale digit image. 
    \item Fashion-MNIST \cite{paper40}: A 10 classes dataset of $70000$ samples. Each sample is a $28\times28$ grayscale image of a fashion product.
\end{itemize}

Our baselines include models from different clustering categories. For classical clustering,  
K-Means \cite{paper14}, Least Squares Non-negative Matrix Factorization (LSNMF) \cite{paper75}, Gaussian mixture models (GMM) \cite{paper47} and Agglomerative Clustering (AC) \cite{paper74} are selected. For subspace clustering, our baselines include two state-of-the-art subspace models. It is worth to note that most existing subspace algorithms are not scalable enough to handle databases with 70000 samples. Therefore, our choice was limited to Scalable Sparse Subspace Clustering by Orthogonal Matching Pursuit (SSC-OMP) \cite{paper78} and Scalable Elastic Net Subspace Clustering (EnSC) \cite{paper79}. For manifold clustering, we opted for normalized-cut spectral clustering (SC) \cite{paper76} and RBF K-Means\cite{paper77}.
As for deep clustering, our experimental protocol include several approaches: JULE \cite{paper22}, DEC \cite{paper27}, IDEC \cite{paper28}, DEPICT \cite{paper36}, DCN \cite{paper29}, VaDE \cite{paper35}, SR-K-Means\cite{paper73} and DeepCluster \cite{paper24}. We have also compared our model against AE+K-Means and AE+FINCH where, the latent samples are clustered using K-Means and FINCH, respectively, after training an autoencoder to reconstruct the data.

\subsubsection{Evaluation Metrics}
ACC \cite{paper41} and NMI \cite{paper42} are the most utilized evaluation metrics in the deep clustering literature. Both of them lie in the range $[0, 1]$. The higher these values are the better our clustering results. Mathematically, ACC and NMI can be defined by Equations \ref{eq:acc} and \ref{eq:nmi}, respectively. In these equations, $y_{true}$ represents the vector of ground-truth labels and $y_{pred}$ is the vector of clustering indexes. $I$ is the mutual information function and $H$ denotes the entropy. Finally, $T$ is the best one-to-one mapping that matches the clustering indexes to their corresponding ground truth labels. This mapping can be found using the Hungarian algorithm \cite{paper43}.

\begin{equation}\label{eq:acc}
    ACC(y_{pred}, \;y_{true}) = max_{T}\left \{ \frac{\sum_{i=1}^{N} 1 \left \{ y_{true}(i)=T(y_{pred}(i)) \right \} }{N}  \right \},
\end{equation}

\begin{equation}\label{eq:nmi}
    NMI(y_{true}, \; y_{true}) = \frac{I(y_{true}, \; y_{pred})}{\frac{1}{2}[H(y_{true}) + H(y_{pred})]}.   
\end{equation}

\subsubsection{Implementation}
For the sake of fair comparison, we opt for the same architecture proposed by a significant number of our deep clustering baselines \cite{paper27, paper28, paper35, paper29}. Our encoder and decoder are fully connected neural networks. The autoencoder has 8 layers with dimensions $d$ - 500 - 500 - 2000 - 10 - 2000 - 500 - 500 - $d$. Except for the bottleneck layer and the last layer, all the other ones rely on ReLu \cite{paper44} activation functions. During the pretraining stage, the autoencoder is trained adversarially end-to-end in competition with a critic network for $13 \times 10^{4}$ iterations. For this phase, all the learnable parameters $\theta_{f}$, $\theta_{g}$ and $\theta_{c}$ are updated using Adam \cite{paper45} with a learning rate equal to $0.0001$ and default values for $\beta_{1}$, $\beta_{2}$, and $\epsilon$ (here we are referring to the optimizer hyperparameters not to the ones of our model). The hyperparameters $\lambda$ and $\alpha$ are set as described in \cite{paper5}. During the clustering phase, the autoencoder is trained until meeting the convergence criterion with a convergence threshold $tol = 1\%$ or reaching a maximal number of iterations $MaxIter = 10^{5}$. The confidence threshold is set to $\kappa = 0.3 \times K$ and its dropping rate is set to $\triangledown \kappa = 0.3 \times \kappa$. The training parameters $\theta_{f}$ and $\theta_{g}$ are updated using SGD with a learning rate $\vartheta = 0.001$ and a momentum equal to $0.9$. Both the pretraining and clustering phases are executed in batches of size $256$ each. DynAE was implemented using Python and Tensorflow \cite{paper46}. The hyperparameters of our baselines are fixed to their default values provided by the authors of the related state-of-the-art works.

\subsection{Results}
Our experimental results can be divided into five sections. In the first section, we compare DynAE with state-of-the-of-art clustering approaches using the standard metrics. In the second section, we show the suitability of our gradual and smooth dynamics in learning unsupervised representations. In the third section, our experiments prove the ability of DynAE to reduce Feature Randomness and Feature Drift using our proposed metrics $\Delta_{FR}$ and $\Delta_{FD}$. In the fourth section, for even fairer comparison, we introduce a balancing hyperparameter $\gamma$, similar to related works \cite{paper27, paper28, paper29, paper36} that can balance between the two losses, to show that DynAE achieves state-of-the-art results for a wide range of $\gamma$ as well. Finally, in the fifth section, we illustrate some qualitative results. In all the following experiments, approaches marked with '*' share the same architecture and the same pretraining weights with DynAE.

\subsubsection{Comparing to state-of-the-art approaches}
In table \ref{table:ACC_NMI_comparison}, we provide a comparative evaluation of our proposed DynAE to several clustering approaches from different categories. Our first evaluation aims to compare DynAE to the different approaches, in terms of clustering effectiveness, using the ACC and NMI metrics. First, we notice that the classical clustering approaches generally outperform the subspace methods. Although subspace clustering is designed to deal with high-dimensional data, it is well known that clusters from high-semantic datasets (e.g., images, sound, text etc) lie on non-linearly shaped manifolds. Subspace methods surmise that the clusters lie on a union of linear subspaces. This assumption contradicts the manifold hypothesis. Thus, it can be a possible explanation to the first observation. As a second observation, it is clear that AC has better ACC and NMI than the other classical methods. In fact, since high-semantic features are hierarchical, merging pairs of clusters gradually can probably help to find high-semantic similarities. Except for AC, we can see that the manifold clustering methods outperform the classical ones on all datasets. Furthermore, we can see from Table \ref{table:ACC_NMI_comparison} that deep clustering methods outperform the other clustering methods significantly. This result confirms the convenience of processing high-dimensional, high-semantic, and large-scale data using deep neural networks. Comparing among the deep clustering strategies, DynAE beats its deep baselines on every dataset by a considerable margin. For example, we can observe that the ACC provided by DynAE is at least 2\% higher than the ACC provided by DEPICT. Besides, the NMI provided by DynAE is at least 5\% higher than the NMI provided by DEPICT. It is important to note that DEC and DEPICT have the same clustering strategy. The main difference between them is their neural architectures. More precisely,  DEC leverages a simple fully-connected architecture, DEPICT makes use of a convolutional architecture.
                                                            
\begin{table}[h!]
  \caption{The ACC and NMI of different clustering approaches. Each category is separated from the other ones by a double horizontal line. {-} indicates that the program ran out of memory. Best method in bold, second best emphasized.}
  \vskip 0.01in 
  \begin{center}
  \begin{small}
  \begin{tabular}{|p{3cm}|c|c|c|c|c|c|c|c|c|c|}
    \hline
    {\textbf{Method}} & \multicolumn{2}{c|}{\textbf{MNIST-full}} & \multicolumn{2}{c|}{\textbf{MNIST-test}} & \multicolumn{2}{c|}{\textbf{USPS}} & \multicolumn{2}{c|}{\textbf{Fashion-MNIST}}\\
    \cline{2-9}
    & \textbf{ACC} & \textbf{NMI} & \textbf{ACC} & \textbf{NMI} & \textbf{ACC} & \textbf{NMI} & \textbf{ACC} & \textbf{NMI} \\ \hline
    \textbf{K-Means} & 0.532 & 0.500 & 0.546 & 0.501 & 0.668 & 0.627 & 0.474 & 0.512 \\ \hline
    \textbf{GMM} & 0.433 & 0.366 & 0.540 & 0.493 & 0.551 & 0.530 & 0.556 & 0.557 \\ \hline
    \textbf{LSNMF} & 0.540 & 0.455 & 0.550 & 0.463  & 0.575 & 0.551 & 0.549 & 0.523 \\ \hline 
    \textbf{AC} & 0.621 & 0.682 & 0.695 & 0.711 & 0.683 & 0.725 & 0.500 & 0.564 \\ \hline \hline
    \textbf{SSC-OMP} & 0.309 & 0.315 & 0.413 & 0.450 & 0.477 & 0.503 & 0.100 & 0.007 \\ \hline
    \textbf{EnSC} & 0.111 & 0.014 & 0.603 & 0.591 & 0.610 & 0.684 & \textbf{0.629} & \textit{0.636} \\ \hline \hline
    \textbf{SC} & 0.656 & 0.731 & 0.660 & 0.704 & 0.649 & 0.794 & 0.508 & 0.575  \\ \hline
    \textbf{RBF K-Means} & {-} & {-} & 0.560 & 0.523 & 0.629 & 0.631 & {-} & {-}  \\ \hline \hline
    \textbf{AE + K-Means} & 0.807 & 0.730 & 0.702 & 0.617 & 0.720 & 0.698 & 0.585 & 0.614\\ \hline
    \textbf{AE + FINCH} & {-} & {-} & 0.709 & 0.754 & 0.704 & 0.788 & {-} & {-}  \\ \hline 
    \textbf{DeepCluster} & 0.797 & 0.661 & 0.854 & 0.713 & 0.562 & 0.540 & 0.542  & 0.510  \\ \hline
    \textbf{DCN} & 0.830 & 0.810 & 0.802 & 0.786 & 0.688 & 0.683  & 0.501 & 0.558  \\ \hline 
    \textbf{DEC} & 0.863 & 0.834 & 0.856 & 0.830 & 0.762 & 0.767 & 0.518 & 0.546 \\ \hline
    \textbf{IDEC} & 0.881 & 0.867 & 0.846 & 0.802 & 0.761 & 0.785 & 0.529 & 0.557 \\ \hline
    \textbf{SR-KMeans} & 0.939 & 0.866 & 0.863 & 0.873 & 0.901 & 0.912 & 0.507 & 0.548 \\ \hline
    \textbf{VaDE} & 0.945 & 0.876 & 0.287 & 0.287 & 0.566 & 0.512 & 0.578 & 0.630  \\ \hline
    \textbf{JULE} & 0.964 & 0.913 & 0.961 & 0.915 & \textit{0.950} & \textit{0.913} &  0.563 & 0.608  \\ \hline
    \textbf{DEPICT} & \textit{0.965} & \textit{0.917} & \textit{0.963} & \textit{0.915} & 0.899 & 0.906 &  0.392 & 0.392 \\ \hline
    \textbf{DynAE} & \textbf{0.987} & \textbf{0.964} & \textbf{0.987} & \textbf{0.963} & \textbf{0.981} & \textbf{0.948} & \textit{0.591} & \textbf{0.642} \\ \hline 
  \end{tabular}
  \label{table:ACC_NMI_comparison}
  \end{small}
  \end{center}
  \vskip 0.1in
\end{table}

Generally speaking, each one of the compared models in Table \ref{table:ACC_NMI_comparison} differs from the other ones in different aspects. Among these aspects: (1) the used architecture, (2) the integrated prior knowledge (e.g., invariance of the samples to small linear geometric transformations), (3) the selected objective function for the pretraining phase and (4) the learning dynamics. In order to show the importance of our learning dynamics, we should neutralize the other aspects. Therefore, we have reimplemented two famous deep clustering models, namely, DEC and IDEC, according to our architecture, integrated prior knowledge, and pretraining objective function. We denote the new obtained models DEC* and IDEC*. It is worth to note that DynAE already shares the same architecture with DEC* and IDEC*. Therefore, we should only pretrain these models to minimize an adversarially constrained interpolation loss regularized with data augmentation. In Table \ref{table:fair_comparison}, we illustrate our obtained results. Based on these results, we make the following observations. First, DEC* and IDEC* outperform their standard versions DEC and IDEC, respectively, by a significant margin. This result proves the importance of our pretraining strategy compared with the vanilla reconstruction performed by DEC and IDEC. Second, we observe that DynAE still outperforms DEC* and IDEC* on all datasets, in terms of ACC and NMI. This result shows the importance of our smooth dynamic objective function in learning discriminative unsupervised representations. 

\begin{table}[h!]
\vskip 0.1in
  \caption{The ACC and NMI of DEC*, IDEC* and DynAE. Best method in bold, second best emphasized.}
  \vskip 0.1in
  \begin{center}
  \begin{small}
  \begin{tabular}{|p{2.7cm}|c|c|c|c|c|c|c|c|}
    \hline
    {\textbf{Method}} & \multicolumn{2}{c|}{\textbf{MNIST-full}} & \multicolumn{2}{c|}{\textbf{MNIST-test}} & \multicolumn{2}{c|}{\textbf{USPS}} & \multicolumn{2}{c|}{\textbf{Fashion-MNIST}} \\
    \cline{2-9}
    & \textbf{ACC} & \textbf{NMI} & \textbf{ACC} & \textbf{NMI} & \textbf{ACC} & \textbf{NMI} & \textbf{ACC} & \textbf{NMI} \\ \hline
    \textbf{DEC*} & 0.971 & 0.929 & 0.968 & 0.920 & 0.963 & 0.910 & 0.575 & 0.589 \\ \hline
    \textbf{IDEC*} & \textit{0.982} & \textit{0.952} & \textit{0.978} & \textit{0.944} & \textit{0.980} & \textit{0.946} & \textit{0.575} & \textit{0.631} \\ \hline
    \textbf{DynAE} & \textbf{0.987} & \textbf{0.964} & \textbf{0.987} & \textbf{0.963} & \textbf{0.981} & \textbf{0.948} & \textbf{0.591} & \textbf{0.642} \\ \hline 
  \end{tabular}
  \label{table:fair_comparison}
  \end{small}
  \end{center}
  \vskip 0.1in
\end{table}

In Table \ref{table:exec_time}, we compare DynAE to several deep clustering methods, in terms of run-time. In this comparison, we exclude the other clustering categories because of their less competitive results in Table \ref{table:ACC_NMI_comparison}. Based on our comparison, we make the following observations. First, we can notice that the execution time of our model is higher than the execution time of DeepCluster, DCN, DEC, IDEC. Even so, the run-time of DynAE is on an equal footing with the ones of DEPICT, SR-Kmeans, and JULE on some datasets. Compared to VaDE, our model has a lower execution time on all evaluated datasets.

\begin{table}[h!]
  \caption{The execution time (in seconds) of different deep clustering approaches.}
  \vskip 0.1in
  \begin{center}
  \begin{small}
  \begin{tabular}{|p{2.7cm}|c|c|c|c|c|c|}
    \hline
    {\textbf{Method}} & {\textbf{MNIST-full}} & {\textbf{MNIST-test}} & {\textbf{USPS}} &{\textbf{Fashion-MNIST}} \\ \hline
    \textbf{DeepCluster} & 1375 & 74 & 64 & 1250 \\ \hline
    \textbf{DCN} & 640 & 55 & 49 & 732 \\ \hline
    \textbf{DEC} & 693 & 58 & 53 & 2384 \\ \hline
    \textbf{IDEC} & 890 & 349 & 110 & 857 \\ \hline
    \textbf{SR-KMeans} & 14872 & 1657 & 1655 & 4551 \\ \hline
    \textbf{VaDE} & 123000 & 15000 & 13000 & 120000 \\ \hline
    \textbf{JULE} & 12500 & 3247 & 2540 & 13100 \\ \hline
    \textbf{DEPICT} & 9561 & 2320 & 1778 & 8581 \\ \hline
    \textbf{DynAE} & 10808 & 9924 & 7910 & 10508 \\ \hline 
  \end{tabular}
  \label{table:exec_time}
  \end{small}
  \end{center}
  \vskip 0.1in
\end{table}

In order to better evaluate the efficiency of our model, we compare its execution time with the ones of DEC* and IDEC*. Based on Table \ref{table:fair_exec_time}, we make the following observations. First, the run-time of DEC and IDEC is significantly lower than the run-time of DEC* and IDEC*, respectively. Hence, we can conclude that the adversarial pretraining phase causes overhead. Second, we observe that the run-time of DynAE is on an equal footing with DEC* and IDEC* on all datasets. Thus, we can conclude that our dynamic clustering phase does not lead to any considerable increase in the execution time.

\begin{table}[h!]
  \caption{The execution time (in seconds) of DEC*, IDEC* and DynAE.}
  \vskip 0.1in
  \begin{center}
  \begin{small}
  \begin{tabular}{|p{2.2cm}|c|c|c|c|}
    \hline
    {\textbf{Method}} & {\textbf{MNIST-full}} & {\textbf{MNIST-test}} & {\textbf{USPS}} &{\textbf{Fashion-MNIST}}  \\ \hline
    \textbf{DEC*} & 9667 & 9092 & 7692 & 10840 \\ \hline
    \textbf{IDEC*} & 9556 & 9160 & 7693 & 9623 \\ \hline
    \textbf{DynAE} & 10808 & 9924 & 7910 & 10508 \\ \hline 
  \end{tabular}
  \label{table:fair_exec_time}
  \end{small}
  \end{center}
  \vskip 0.1in
\end{table}

\subsubsection{Dynamic learning}
\label{Sec:Dynamic learning}
In this section and the following two sections, we show results of a single dataset, namely, MNIST. The reader can refer to the supplementary material (\ref{MNIST-test}, \ref{USPS} and \ref{Fashion-MNIST}) to find results on other datasets. It is worth to note that the same conclusions can be drawn using MNIST-full or MNIST-test or USPS or Fashion-MNIST. In this part of our experimental protocol, we aim to show the effect of training using a smooth dynamic loss function. 

\begin{figure*}[ht]
\vskip 0.2in
\centering
    \subfigure[Clustering loss.]{\includegraphics[width=0.45\linewidth]{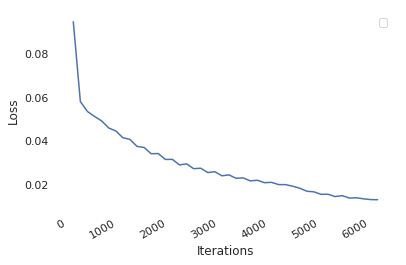}}
    \subfigure[Dynamics of the clustering loss.]{\includegraphics[width=0.45\linewidth]{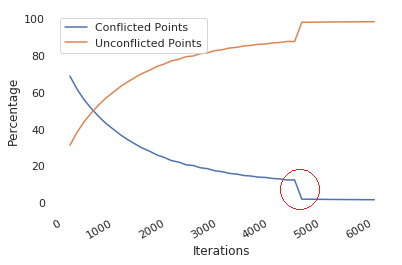}}
    \caption{Learning dynamics during training on MNIST.}
\label{fig:loss_nb_MNIST}
\end{figure*}

Figure \ref{fig:loss_nb_MNIST} illustrates the learning dynamics during training on MNIST. We can describe the dynamics of our learning system by the evolution of the optimized objective function, which in turn can be described by the amount of reconstruction $\tau$. In Figure \ref{fig:loss_nb_MNIST}.a, we can clearly notice that the clustering loss decreases almost smoothly. Hence, we can conclude that the gradual change of our objective function does not perturb the learning process. In Figure \ref{fig:loss_nb_MNIST}.b, we draw the percentage of conflicted samples (respectively the percentage of unconflicted samples) against the number of iterations. At every iteration, the percentage of conflicted points represents the amount of reconstruction $\tau$. As we can see from Figure \ref{fig:loss_nb_MNIST}.b, the percentage of conflicted points decreases smoothly until reaching the red circle. Close to iteration $5000$, the dropping rate of the conflicted samples starts to slow down. At this stage, the model has reached local stability. In order to avoid the local optimum, the centroids, $\beta_{1}$ and $\beta_{2}$ are updated automatically at this stage. This update causes an abrupt decrease in the number of conflicted samples, as we can see inside the red circle. Although the centroids, $\beta_{1}$, and $\beta_{2}$ are never updated before the red circle, the number of conflicted samples decreases smoothly from one iteration to another. This result confirms that the knowledge acquired from clustering the most reliable samples gradually makes more difficult samples (conflicted) become reliable for clustering (unconflicted).


\begin{figure*}[ht]
\vskip 0.2in
\centering
    \subfigure[Unconflicted data.]{\includegraphics[width=0.45\linewidth]{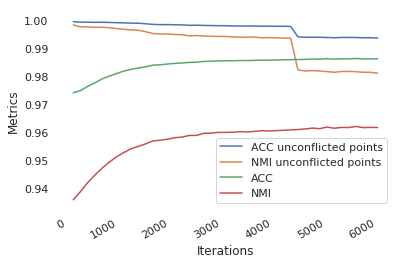}}
    \subfigure[Conflicted data.]{\includegraphics[width=0.45\linewidth]{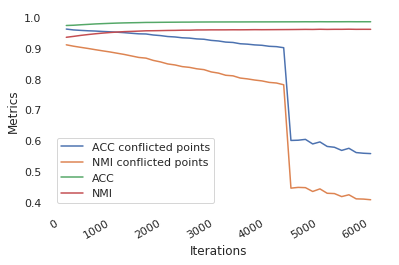}} 
    \caption{ACC and NMI values during training on MNIST.}
\label{fig:ACC_NMI_MNIST}
\end{figure*}

Figure \ref{fig:ACC_NMI_MNIST} illustrates the evolution of ACC and NMI during training on MNIST. First, we observe that both metrics increase smoothly as computed using the whole training set. In Figure \ref{fig:ACC_NMI_MNIST}.a, we observe that the ACC and NMI of the unconflicted samples decrease slowly before reaching local stability between the iterations 4000 and 5000. This result is expected. In fact, as long as the number of reliable samples increases, there is a low probability of making some errors. These errors are associated with training using the newly clustered samples. Most interestingly, we observe that the accuracy of the unconflicted subset is greater than 99.8\% at iteration 4000. At this stage, the size of the unconflicted subset constitutes more than 80\% of the whole training set. In Figure \ref{fig:ACC_NMI_MNIST}.b, we observe that the ACC and NMI of the conflicted samples decrease rapidly.  As the training progresses, the remaining conflicted points are more challenging than the newly clustered ones. Therefore, their ACC and NMI are gradually decreasing.

\subsubsection{Feature Randomness and Feature Drift}
\label{Sec:Feature Randomness and Feature Drift}
To show the ability of our model to reduce Feature Randomness and Feature Drift, we propose to ablate the dynamic mechanism and compare the obtained model with DynAE. We denote the model obtained after performing ablation by DC*. DC* has the same optimizer and pretraining phase as DynAE. However, it is trained to optimize joint embedded K-Means and reconstruction. We also compare DynAE with IDEC* and ADEC.

\begin{figure*}[ht]
\vskip 0.2in
\centering
    \subfigure[DynAE.]{\includegraphics[width=0.45\linewidth]{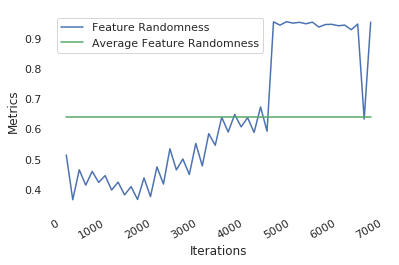}}
    \subfigure[ADEC.]{\includegraphics[width=0.45\linewidth]{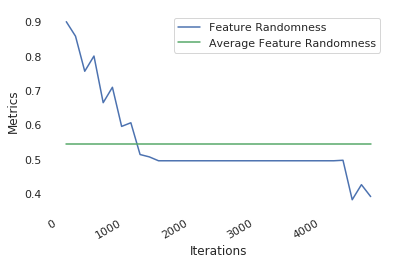}} 
    \subfigure[DC*.]{\includegraphics[width=0.45\linewidth]{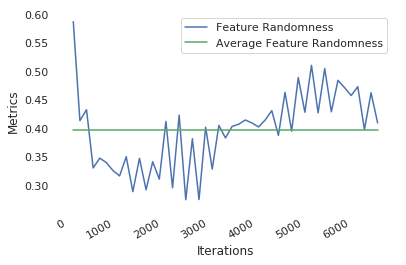}} 
    \subfigure[IDEC*.]{\includegraphics[width=0.45\linewidth]{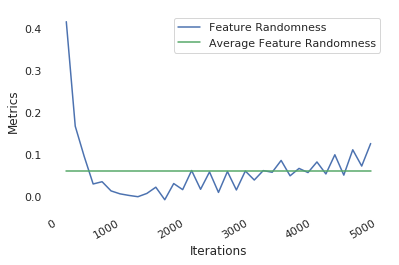}} 
    \caption{$\Delta_{FR}$ values during training on MNIST.}
\label{fig:Delta_FR_MNIST}
\end{figure*}

Figure \ref{fig:Delta_FR_MNIST} illustrates the evolution of $\Delta_{FR}$ values for DynAE, ADEC, DC*, and IDEC* during training on MNIST. From this figure, we can see that the average $\Delta_{FR}$ for DynAE is higher than the average $\Delta_{FR}$ for ADEC and significantly higher than the average $\Delta_{FR}$ for DC* and IDEC*. At the beginning of the training process, we observe that the gradient of ADEC is a good approximation to the supervised gradient ($\Delta_{FR} = 0.9$). As the training progresses, $\Delta_{FR}$ for DynAE tends to increase with some fluctuations. However, $\Delta_{FR}$ for ADEC decreases and  $\Delta_{FR}$ for DC* and IDEC* does not seem to have a clear increasing tendency. At the end of the training process, $\Delta_{FR}$ for DynAE reaches a value greater than $0.9$. This means that the direction of the gradient of DynAE becomes very close to the direction of the supervised gradient. Hence, we can confirm that our dynamic learning mechanism is suitable for mitigating the effect of Feature Randomness.

If we compare the learning curves of DynAE using both metrics $ACC$ and $\Delta_{FR}$, we observe that the learning curve evaluated based on $ACC$ increases smoothly without any fluctuations. As opposed to that, the learning curve of DynAE evaluated based on $\Delta_{FR}$ oscillates. This result suggests that the evolution of $ACC$ hide some important information related to the learning process.  





\begin{figure*}[ht]
\vskip 0.2in
\centering
    \subfigure[DynAE.]{\includegraphics[width=0.45\linewidth]{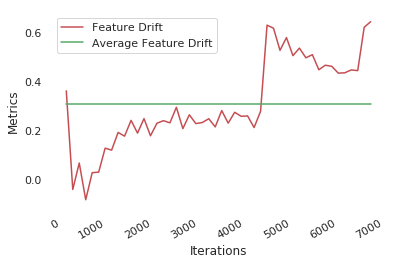}}
    \subfigure[ADEC.]{\includegraphics[width=0.45\linewidth]{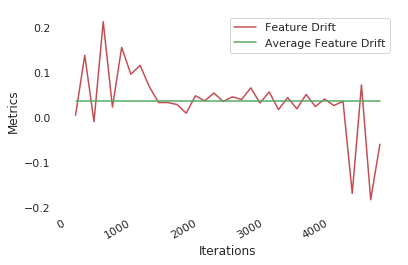}} 
    \subfigure[DC*.]{\includegraphics[width=0.45\linewidth]{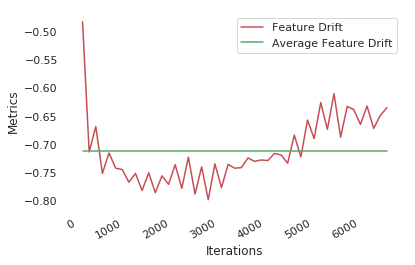}} 
    \subfigure[IDEC*.]{\includegraphics[width=0.45\linewidth]{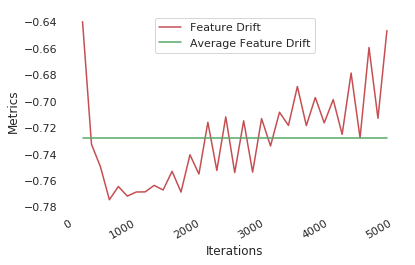}} 
    \caption{$\Delta_{FD}$ values during training on MNIST.}
\label{fig:Delta_FD_MNIST}
\end{figure*}

Figure \ref{fig:Delta_FD_MNIST} illustrates the evolution of $\Delta_{FD}$ for DynAE, ADEC, DC*, and IDEC* during training on MNIST. From this figure, we observe that the average $\Delta_{FD}$ for DynAE is higher than the average $\Delta_{FD}$ for ADEC and considerably higher than the average $\Delta_{FD}$ for DC* and IDEC*. Furthermore, $\Delta_{FD}$ is always negative for DC* and IDEC*, with an average value around $-0.7$. This negative value reflects the strong competition between the gradient of the vanilla reconstruction and the gradient of the embedded clustering loss. Based on these observations, we can confirm that our dynamic learning mechanism is suitable for alleviating the effect of Feature Drift by gradually eliminating the reconstruction loss.

\subsubsection{Effect of introducing a balancing hyperparameter}
\label{Sec:Balancing hyperparameter}
\begin{figure*}[ht]
\vskip 0.2in
\centering
    \subfigure[DynAE.]{\includegraphics[width=0.45\linewidth]{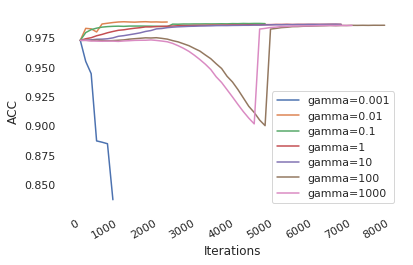}}
    \subfigure[IDEC*.]{\includegraphics[width=0.45\linewidth]{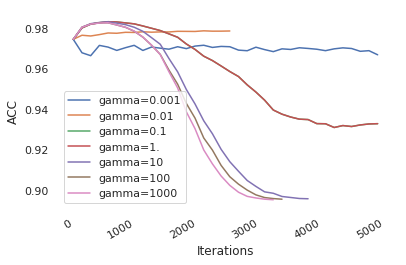}} 
    \caption{Effect of the balancing hyperparameter during training on MNIST.}
\label{fig:balancing_hyperparameter_MNIST}
\end{figure*}

In order to evaluate the impact of a balancing hyperparameter $\gamma$ on DynAE (similar to IDEC), we propose to change the objective function of our model to $L = L_{1} + \gamma L_{2}$. Figure \ref{fig:balancing_hyperparameter_MNIST} illustrates the effect of $\gamma$ on the training of DynAE and IDEC*. In this experiment, we tried different $\gamma$ values from the set $\left\{ 10^{-3}, \: 10^{-2}, \: 10^{-1}, \: 1, \: 10, \: 10^{2}, \: 10^{3}\right\}$. We found $\gamma = 0.01$ provides us the best learning curves for both models. For IDEC*, apart from  $\gamma = 0.01$, all the other values make the learning curves detoriate drastically. Hence, we conclude that IDEC* is very sensitive to the setting of $\gamma$. However, DynAE almost reaches the same ACC at the end of the training process for 6 $\gamma$ values out of 7.
For $\gamma = 0.001$, $\gamma L_{2}$ becomes significantly smaller compared to $L_{1}$. Thus, the gradient of the self-supervised loss  drifts the features learned in the direction of the embedded clustering loss $\gamma L_{2}$. For this reason, the learning curve of DynAE, for $\gamma = 0.001$, has a collapsing shape. As $\gamma$ grows, $\gamma L_{2}$ becomes significantly greater than $L_{1}$. Hence, pseudo-supervision becomes dominant of the training process, leading to Feature Randomness. Going from $0.01$ to $1000$, we observe that the number of iterations required for convergence increases. Added to that, the final ACC of DynAE slightly decreases when $\gamma$ increases. However, as we can see from the figure, it is apparent that DynAE is considerably less sensitive to the setting of $\gamma$ when compared to IDEC*. This result confirms that our model offers a better trade-off between Feature Randomness and Feature Drift compared to \textit{IDEC*}.
 
\begin{figure}
\vskip 0.2in
\begin{center}
\centerline{\includegraphics[width=250pt, height=160pt]{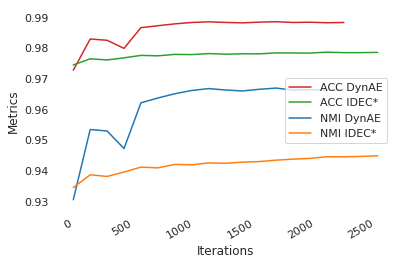}}
\caption{ACC and NMI during training on MNIST ($\gamma = 0.01$).}
\label{fig:ACC_NMI_tunned_mnist}
\end{center}
\vskip -0.2in
\end{figure}

\begin{figure}
\vskip 0.2in
\centering
    \subfigure[DynAE.]{\includegraphics[width=180pt, height=180pt]{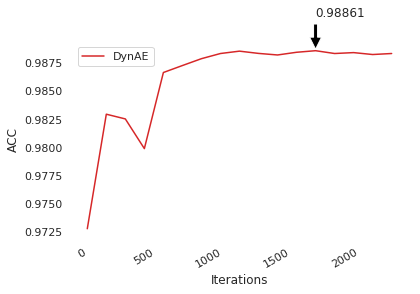}}
    \subfigure[IDEC*.]{\includegraphics[width=180pt, height=180pt]{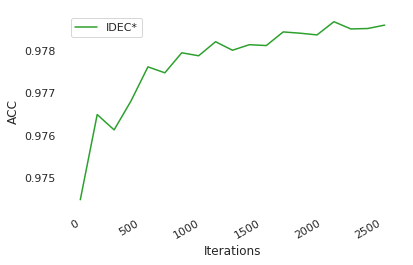}} 
    \caption{ACC values during training on MNIST ($\gamma = 0.01$).}
\vskip -0.2in
\label{fig:ACC_tuned_MNIST}
\end{figure}

\begin{figure}
\vskip 0.2in
\centering
    \subfigure[DynAE.]{\includegraphics[width=180pt, height=180pt]{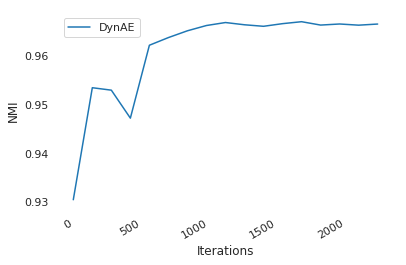}}
    \subfigure[IDEC*.]{\includegraphics[width=180pt, height=180pt]{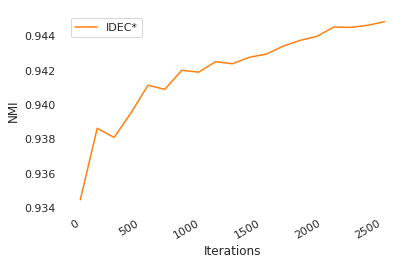}} 
    \caption{NMI values during training on MNIST ($\gamma = 0.01$).}
\vskip -0.2in
\label{fig:NMI_tuned_MNIST}
\end{figure}

We further evaluate the impact of Feature Drift on ACC and NMI with $\gamma = 0.01$, which provides the best learning curves for both DynAE and IDEC*. In Figure \ref{fig:ACC_NMI_tunned_mnist}, we draw the evolution of ACC and NMI for DynAE and IDEC* during training on MNIST with $\gamma = 0.01$. In Figure \ref{fig:NMI_tuned_MNIST} and \ref{fig:ACC_tuned_MNIST}, we zoom in to each curve separately to inspect better. We notice that the ACC and NMI curves of DynAE are nearly smoothly increasing during the training process. However, the ACC and NMI curves of IDEC* undergo noticeable fluctuations. A possible explanation of these fluctuations is the strong competition between embedded clustering and vanilla reconstruction, which is reduced in DynAE. We also see that the highest ACC value for DynAE during training on MNIST is $0.98861$. Thus, it is likely possible to improve results reported in Table \ref{table:ACC_NMI_comparison} further by tuning the balancing hyperparameter. However, the reported balancing hyperparameter-free DynAE (i.e., $L = L_{1} + L_{2}$) achieves competitive results without the requirement of tuning due to the weaker competition between embedded clustering and vanilla reconstruction.

\subsubsection{Qualitative results}
\begin{figure*}[ht]
\vskip 0.2in
\centering
    \subfigure[MNIST-full.]{\includegraphics[width=0.23\linewidth]{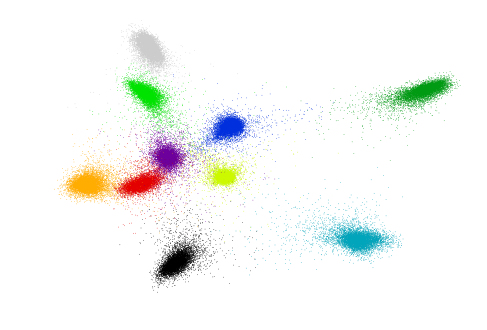}} 
    \subfigure[MNIST-test.]{\includegraphics[width=0.23\linewidth]{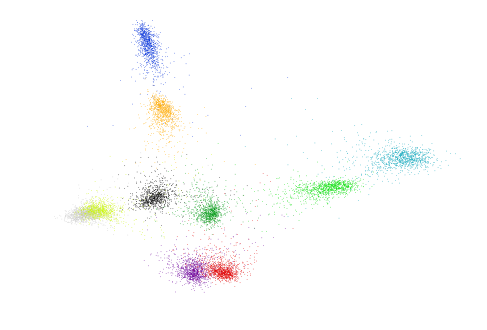}}
    \subfigure[USPS.]{\includegraphics[width=0.23\linewidth]{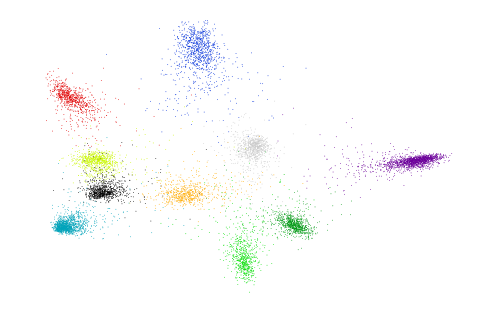}} 
    \subfigure[Fashion-MNIST.]{\includegraphics[width=0.23\linewidth]{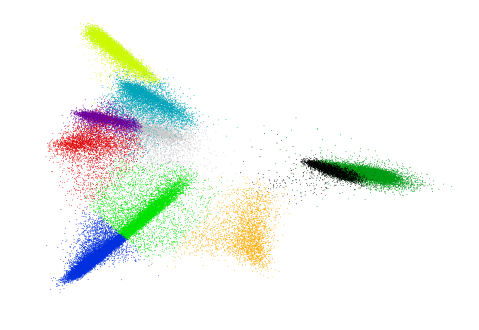}}
    \caption{Visualizing 2D embedded subspaces to show the discriminative ability of DynAE.}
    \label{fig:Visualization of embedding subspaces}
\vskip -0.2in
\end{figure*}

\begin{figure*}[ht]
\vskip 0.2in
\centering
    \subfigure[MNIST.]{\includegraphics[width=0.45\linewidth]{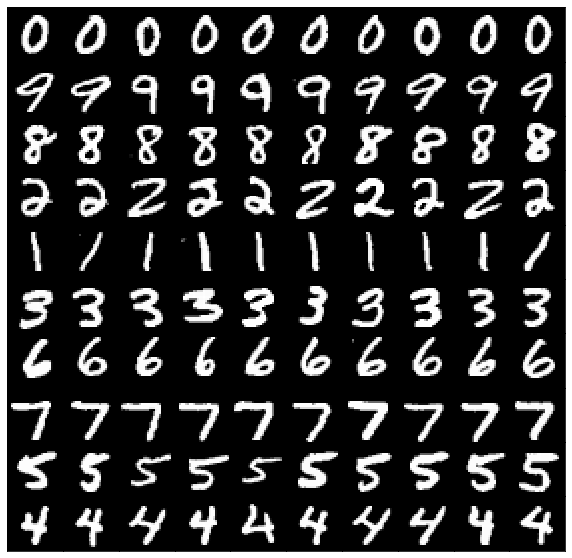}} 
    \subfigure[Fashion MNIST.]{\includegraphics[width=0.45\linewidth]{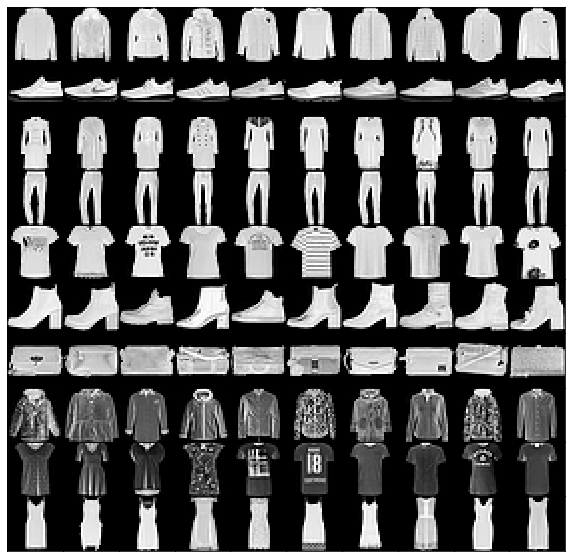}}
    \caption{Top 10 high-confidence images from each cluster displayed in each row.}
\vskip -0.2in
\label{fig:top_cluster_images}
\end{figure*}

The discriminative ability of DynAE can be illustrated in Figure \ref{fig:Visualization of embedding subspaces}. In this figure, we visualize 2D embedded representations for each dataset. These representations are obtained by applying PCA to the latent representations at the end of the clustering phase. Based on our results, we can see that the clusters are well-separated for all datasets. In Figure \ref{fig:top_cluster_images}, the top 10 high-confidence images from each cluster are displayed in each row. As we can notice, DynAE has managed to capture semantic similarities.

\section{Conclusion}
In this article, we have introduced Dynamic Autoencoder, a deep clustering model that integrates smooth learning dynamics. Our proposition consists of gradually and smoothly transforming a self-supervised objective into a pseudo-supervised one. Empirical studies show the effectiveness and suitability of our model in clustering benchmark datasets. In terms of ACC and NMI, DynAE achieves state-of-the-art results compared to the most relevant deep clustering methods. Added to that, experimental evaluations have shown that our approach offers a better trade-off between Feature Randomness and Feature Drift. We strongly believe that the simple but intuitive formulation of DynAE has a lot more potential. In future work, we would like to try a more sophisticated architecture (e.g., VGG, AlexNet, and ResNet32) to cluster larger datasets with higher semanticity. It is also interesting to investigate other techniques, apart from K-Means, for computing the embedded centroids. While jointly performing pseudo-supervision and self-supervision is of paramount importance, it is still unclear if the linear combination used by state-of-the-art autoencoder-based models is the best formalism. Studying other possible formulations can be an interesting line of research.

\bibliographystyle{unsrt}
\bibliography{references.bib}

\newpage
\appendix

\section{Experimental Results on MNIST-test}
\label{MNIST-test}

\begin{figure*}[!ht]
\vskip 0.2in
\centering
    \subfigure[Clustering loss.]{\includegraphics[width=200pt, height=200pt]{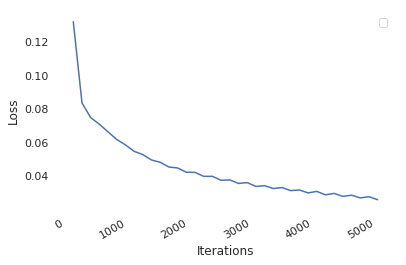}} 
    \subfigure[Dynamics of the clustering loss.]{\includegraphics[width=200pt, height=200pt]{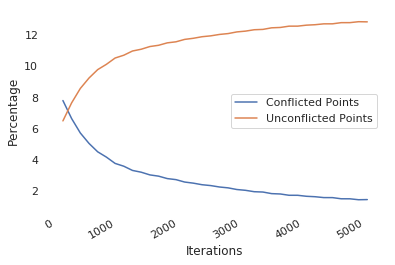}}
    \caption{Learning dynamics during training on MNIST-test.}
\label{fig:loss_nb_MNIST-TEST}
\vskip 0.2in
\end{figure*}

\begin{figure*}[!ht]
\vskip 0.2in
\centering
    \subfigure[Unconflicted data.]{\includegraphics[width=200pt, height=200pt]{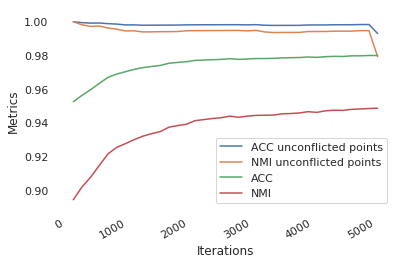}}
    \subfigure[Conflicted data.]{\includegraphics[width=200pt, height=200pt]{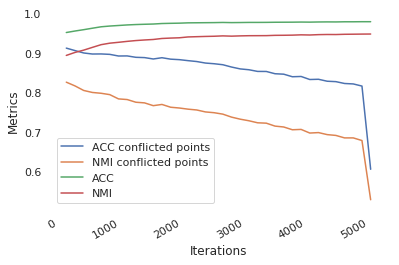}} 
    \caption{ACC and NMI values during training on MNIST-test.}
\label{fig:ACC_NMI_MNIST-TEST}
\vskip 0.2in
\end{figure*}

\begin{figure*}[!ht]
\vskip 0.2in
\centering
    \subfigure[DynAE.]{\includegraphics[width=210pt, height=260pt]{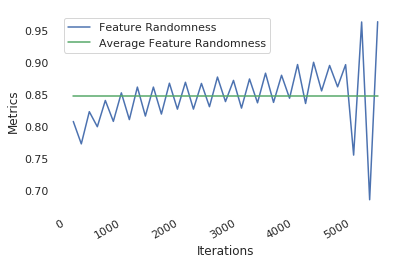}}
    \subfigure[DC*.]{\includegraphics[width=210pt, height=260pt]{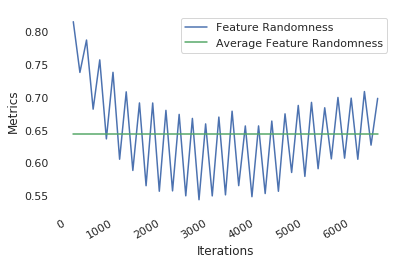}}
    \subfigure[IDEC*.]{\includegraphics[width=210pt, height=260pt]{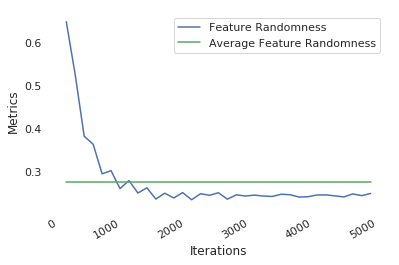}} 
    \caption{$\Delta_{FR}$ values during training on MNIST-test.}
\label{fig:Delta_FR_MNIST-TEST}
\vskip 0.3in
\end{figure*}

\begin{figure*}[!ht]
\vskip 0.2in
\centering
    \subfigure[DynAE.]{\includegraphics[width=210pt, height=260pt]{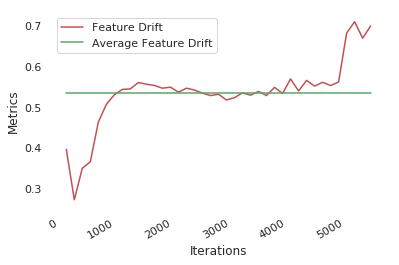}}
    \subfigure[DC*.]{\includegraphics[width=210pt, height=260pt]{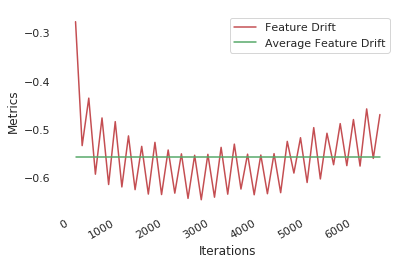}}
    \subfigure[IDEC*.]{\includegraphics[width=210pt, height=260pt]{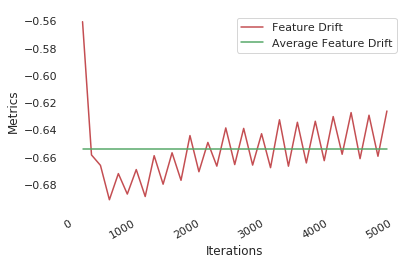}} 
    \caption{$\Delta_{FD}$ values during training on MNIST-test.}
\label{fig:Delta_FD_MNIST-TEST}
\vskip 0.2in
\end{figure*}

\newpage 

\section{Experimental Results on USPS}
\label{USPS}

\begin{figure*}[!ht]
\vskip 0.2in
\centering
    \subfigure[Clustering loss.]{\includegraphics[width=200pt, height=200pt]{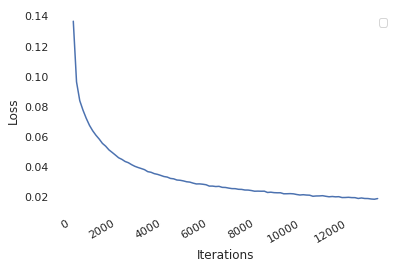}} 
    \subfigure[Dynamics of the clustering loss.]{\includegraphics[width=200pt, height=200pt]{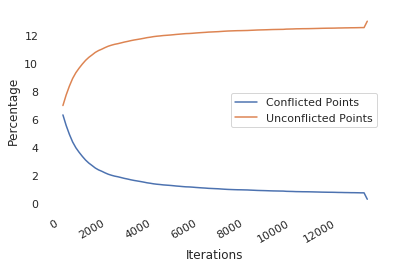}}
    \caption{Learning dynamics during training on USPS.}
\label{fig:loss_nb_USPS}
\vskip 0.2in
\end{figure*}

\begin{figure*}[!ht]
\vskip 0.2in
\centering
    \subfigure[Unconflicted data.]{\includegraphics[width=200pt, height=200pt]{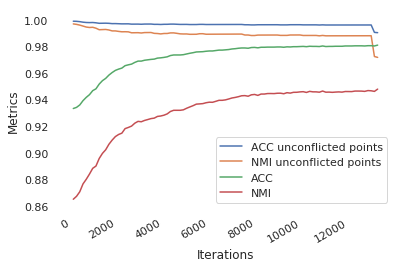}}
    \subfigure[Conflicted data.]{\includegraphics[width=200pt, height=200pt]{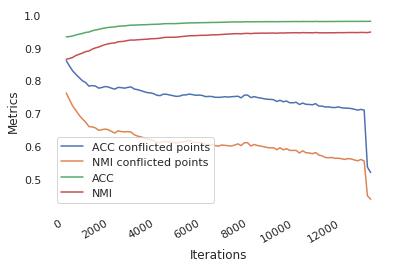}} 
    \caption{ACC and NMI values during training on USPS.}
\label{fig:ACC_NMI_USPS}
\vskip 0.2in
\end{figure*}

\begin{figure*}[!ht]
\vskip 0.2in
\centering
    \subfigure[DynAE.]{\includegraphics[width=210pt, height=260pt]{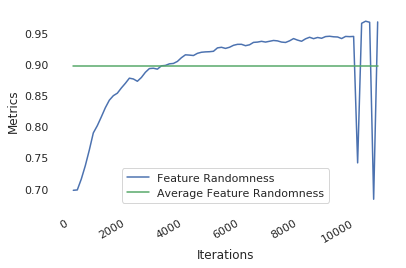}}
    \subfigure[DC*.]{\includegraphics[width=210pt, height=260pt]{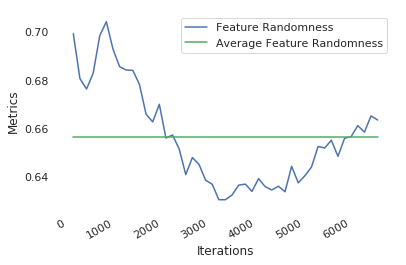}} 
    \subfigure[IDEC*.]{\includegraphics[width=210pt, height=260pt]{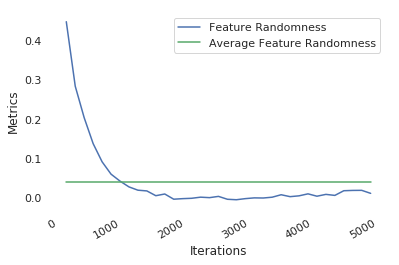}} 
    \caption{$\Delta_{FR}$ values during training on USPS.}
\label{fig:Delta_FR_USPS}
\vskip 0.2in
\end{figure*}

\begin{figure*}[!ht]
\vskip 0.2in
\centering
   \subfigure[DynAE.]{\includegraphics[width=210pt, height=265pt]{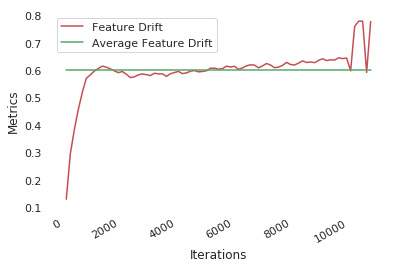}}
   \subfigure[DC*.]{\includegraphics[width=210pt, height=265pt]{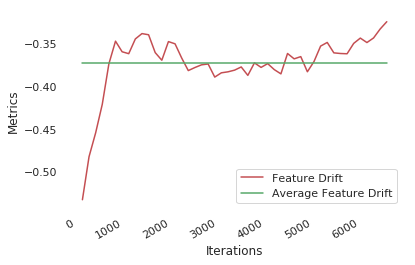}} 
   \subfigure[IDEC*.]{\includegraphics[width=210pt, height=265pt]{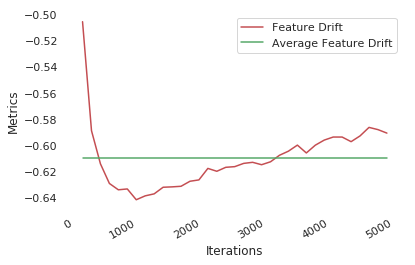}} 
    \caption{$\Delta_{FD}$ values during training on USPS.}
\label{fig:Delta_FD_USPS}
\vskip 0.2in
\end{figure*}

\newpage

\section{Experimental Results on Fashion-MNIST}
\label{Fashion-MNIST}

\begin{figure*}[!ht]
\vskip 0.2in
\centering
    \subfigure[Clustering loss.]{\includegraphics[width=200pt, height=200pt]{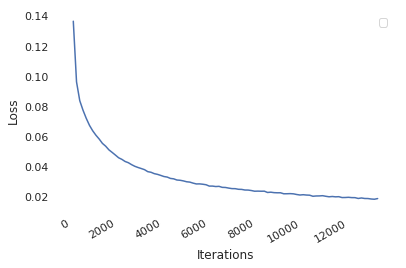}} 
    \subfigure[Dynamics of the clustering loss.]{\includegraphics[width=200pt, height=200pt]{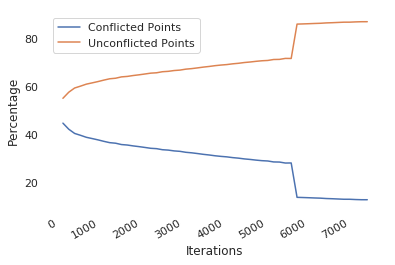}}
    \caption{Learning dynamics during training on FMNIST.}
\vskip -0.2in
\label{fig:loss_nb_FMNIST}
\vskip 0.2in
\end{figure*}

\begin{figure*}[!ht]
\vskip 0.2in
\centering
    \subfigure[Unconflicted data.]{\includegraphics[width=200pt, height=200pt]{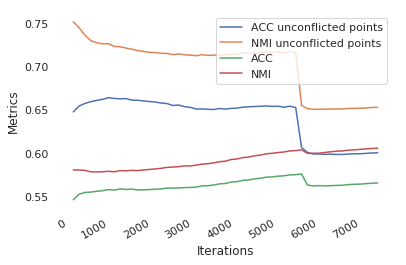}}
    \subfigure[Conflicted data.]{\includegraphics[width=200pt, height=200pt]{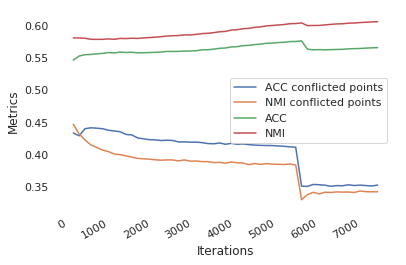}} 
    \caption{ACC and NMI values during training on MNIST.}
\vskip -0.2in
\label{fig:ACC_NMI_FMNIST}
\vskip 0.2in
\end{figure*}

\begin{figure*}[!ht]
\vskip 0.2in
\centering
    \subfigure[DynAE.]{\includegraphics[width=210pt, height=260pt]{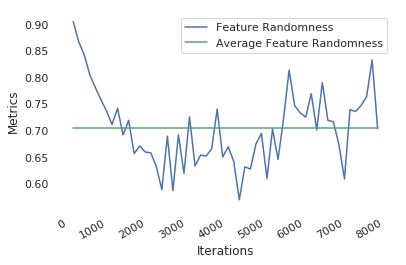}}
    \subfigure[DC*.]{\includegraphics[width=210pt, height=260pt]{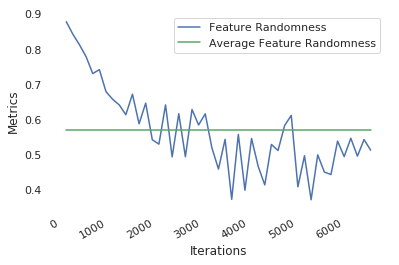}} 
    \subfigure[IDEC*.]{\includegraphics[width=210pt, height=260pt]{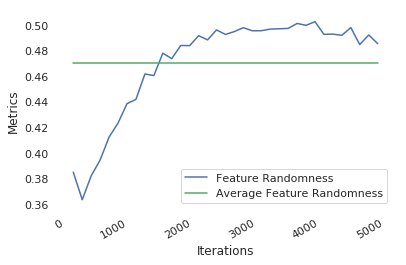}} 
    \caption{$\Delta_{FR}$ values during training on Fashion-MNIST.}
\vskip -0.2in
\label{fig:Delta_FR_FMNIST}
\vskip 0.2in
\end{figure*}

\begin{figure*}[!ht]
\vskip 0.2in
\centering
    \subfigure[DynAE.]{\includegraphics[width=210pt, height=260pt]{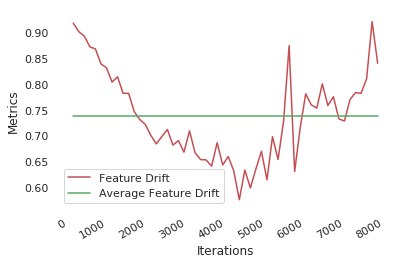}}
    \subfigure[DC*.]{\includegraphics[width=210pt, height=260pt]{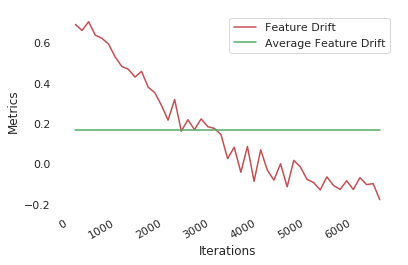}}
    \subfigure[IDEC*.]{\includegraphics[width=210pt, height=260pt]{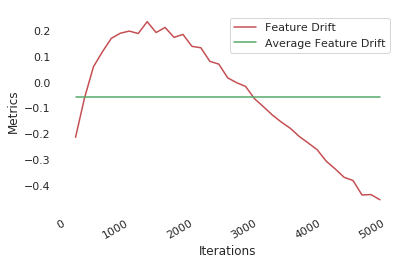}} 
    \caption{$\Delta_{FD}$ values during training on Fashion-MNIST.}
\vskip -0.2in
\label{fig:Delta_FD_FMNIST}
\vskip 0.2in
\end{figure*}

\end{document}